# A COMPARISON OF TWO SMOOTHING METHODS
# FOR WORD BIGRAM MODELS

by

Linda Charlene Bauman Peto

A thesis submitted in conformity with the requirements
for the degree of Master of Science
Graduate Department of Computer Science
University of Toronto

# Abstract

Word bigram models estimated from text corpora require smoothing methods to estimate the probabilities of unseen bigrams. The deleted estimation method uses the formula:

$$\Pr(i \mid j) = \lambda f_i + (1-\lambda) f_{i|j},$$

where $f_i$ and $f_{i|j}$ are the relative frequency of $i$ and the conditional relative frequency of $i$ given $j$, respectively, and $\lambda$ is an optimized parameter. MacKay (1994) proposes a Bayesian approach using Dirichlet priors, which yields a different formula:

$$\Pr(i \mid j) = \left(\frac{\alpha}{F_j + \alpha}\right) m_i + \left(1 - \frac{\alpha}{F_j + \alpha}\right) f_{i|j}$$

where $F_j$ is the count of $j$ and $\alpha$ and $m_i$ are optimized parameters. This thesis describes an experiment in which the two methods were trained on a two-million-word corpus taken from the Canadian *Hansard* and compared on the basis of the experimental perplexity that they assigned to a shared test corpus. The methods proved to be about equally accurate, with MacKay's method using fewer resources.

# Acknowledgements

I would like to thank my supervisors, Graeme Hirst and Dekai Wu. Dekai suggested the topic of this thesis and provided technical guidance, over and above his duties at the Hong Kong University of Science and Technology. Graeme coached me through departmental procedures, corrected the writing style, and exhibited patience and optimism.

David MacKay helpfully explained the details of his method, and advised me regarding the implementation. He also served as the external reader. Bill Gale, Radford Neal, and Peter Brown answered statistical questions.

Finally, I would like to thank my family, James and Michael, for their love and encouragement, and God, who makes all things possible.

# Contents

# 1  Introduction

Today, statistical techniques are finding wide application in natural language processing. They are used in speech processing and optical character recognition to distinguish between similar words, in parsing to disambiguate word senses and phrasal attachments, and in machine translation to choose natural-sounding translation equivalents, among other tasks. In all of these applications, the approach is to develop a statistical model from an existing body of data (the *training corpus*) and then to use that model to interpret new, previously unseen, data.

Since natural language is somewhat unruly, completely accurate statistical models for these tasks would be very complex. In practice, people adopt simplifying assumptions about the statistical models underlying natural language, so that the models can easily be described and worked with mathematically.

One very simple type of model is the $n$-gram model, in which the probability of the next symbol depends only on the $n-1$ symbols preceding it. Trigram models, where the probability of the $y$th symbol depends on the $(y-1)$th and $(y-2)$th symbols, are frequently used, tractable in most cases, and provide reasonable results for many tasks. Bigram models, in which the $y$th symbol depends only on the $(y-1)$th symbol, while less powerful than trigram models, are also simpler mathematically, and are useful for exploring new methods and theories. I use bigram models of word sequences in text as the basis for the comparisons in this thesis.

Once the basic model is chosen, there may be a number of parameters to estimate. In $n$-gram models, the conditional probabilities of each possible symbol given the $n-1$ previous symbols must be found. A simple solution would be to use the conditional relative frequencies of $n$-grams in the training corpus as predictive probabilities. There is a problem with using relative frequencies, however: there might be "holes" in the training data. Suppose a certain $n$-gram never occurs in the training data. Then the conditional relative frequency of that $n$-gram is zero. A predictive probability of zero suggests that the $n$-gram cannot possibly occur, but it is not reasonable to conclude that an $n$-gram is impossible simply because it did not occur in the training data.

One could try to avoid the problem by using enough training data to include every possible legitimate combination. But how much training data is enough? There might even be grammatical combinations of words that have never been uttered yet.

What is needed is a way of assigning tiny positive probabilities to possible sequences of symbols that don't appear in the training data, while maintaining the predictive power and utility of the entire distribution. This is called *smoothing* the distribution. A number of smoothing methods have already been developed for natural language problems. When a new smoothing method is proposed, it is useful to know how it compares with other existing smoothing techniques. The point of this thesis is to compare two existing smoothing methods for word bigram models experimentally.

Different smoothing techniques produce different probability distributions that function as rough approximations of the "true" model underlying language. But since the "true" model, if it exists, is unknown, how can the "goodness" of a smoothing method be measured? In this thesis, I adopt a utilitarian perspective. Remember that these models are being developed in order to identify the mostly likely interpretation of some future data. Therefore, I compare the smoothing methods on the basis of how well the resulting models identify the correct interpretation of some previously unseen data: that is, how probable the correct interpretation is under each model. In the word bigram models used in this thesis, the "interpretation" is simply the word order, and the best model is the one that assigns highest probability to the word orders actually occurring in some previously unseen sentences. A readable explanation of the theoretical basis for this approach to model comparison can be found in MacKay (1992). Of course, in determining the utility of a smoothing method for real applications, it is also necessary to consider the resources required to develop the model, and I will discuss this as well.

## 2 Background

### 2.1 Word Bigram Models

Under a word bigram model, a sentence is generated by first randomly choosing an initial word, and then randomly choosing the next word with probabilities based on the previous word, until a complete sentence is constructed. A text is made up of many sentences so generated. By examining a generated text statistically, one attempts to estimate the conditional probability of each different word pair in the underlying bigram model that generated it.

In this thesis, the training corpus (the generated text) is made up of $n$ sentences labelled $s_1, \ldots, s_n$. Each sentence $s_x$ is made up of $len(s_x)$ tokens labelled $w_{x,1}, \ldots, w_{x,len(s_x)}$. A token can be a word, a punctuation mark, or a special symbol. The first token of each sentence is necessarily $<s>$, and the last token is necessarily $</s>$. The probability of a single sentence in a bigram model is

$$\Pr(s_x) = \Pr(w_{x,1}) \prod_{y=2}^{len(s_x)} \Pr(w_{x,y} \mid w_{x,y-1}). \tag{1}$$

But since $\Pr(w_{x,1} = <s>) = 1$, this simplifies to:

$$\Pr(s_x) = \prod_{y=2}^{len(s_x)} \Pr(w_{x,y} \mid w_{x,y-1}). \tag{2}$$

The probability of the entire training corpus is then:

$$\Pr(s_1, \ldots, s_n) = \prod_{x=1}^{n} \prod_{y=2}^{len(s_x)} \Pr(w_{x,y} \mid w_{x,y-1}). \tag{3}$$

The goal is to find the predictive conditional probability matrix $Q$, where each element $q_{i|j} = \Pr(w_{x,y} = i \mid w_{x,y-1} = j)$. $Q$ is a $W \times W$ matrix, where $W$ is the number of distinct words in the vocabulary. There are an infinite number of possible assignments of values to the $q_{i|j}$, and the goal is to get as close as possible to the "true" value of $Q$ by examining the training data.

Let $F_{i|j}$ be the number of times the bigram $ji$ occurs in the training data, and let $F_j$ be the number of times token $j$ occurs.[1] The conditional relative frequency is

$$f_{i|j} = \frac{F_{i|j}}{F_j}. \tag{4}$$

Setting $q_{i|j} = f_{i|j}$ produces the maximum likelihood estimator of $Q$. As I explained earlier, the maximum likelihood estimator of $Q$ is inadequate because $f_{i|j} = 0$ for any bigram $ji$ that does not occur in the training corpus, and a probability of zero is fatal in most applications. The following smoothing methods address this problem.

## 2.2 Deleted Estimation

Bahl, Jelinek, and Mercer (1983; Jelinek and Mercer 1980) describe a smoothing method for Markov models, which is usually called *deleted estimation*.[2,3] Details of how it works for Markov models can be found in their papers; I present a version for the special case of bigram models here.

Let $i$ and $j$ be two words that appear in the training corpus. With the deleted estimation smoothing method, $q_{i|j}$ is a linear combination of the conditional relative frequency of $i$ given $j$ ($f_{i|j}$) with the relative frequency of $i$ ($f_i$).

$$q_{i|j} = \lambda f_i + (1 - \lambda) f_{i|j}. \tag{5}$$

The proportions of the mixture are governed by a parameter $\lambda$. $\lambda$ is between 0 and 1, and it is estimated from the training data to maximize the predictive power of the resulting distribution. The idea is that, while there might not be sufficient occurrences of bigrams starting with $j$ for the conditional relative frequency to be an accurate estimate of the probability of $i$ occurring after $j$, the relative frequency of $i$ is a good estimator of the probability of $i$ occurring at all. The combination is a neat formula that assigns plausible non-zero probabilities to all bigrams composed of words from the training corpus.

Bahl, Jelinek, and Mercer (1983) actually use several different $\lambda$s depending on the relative frequency of $j$. When there are many occurrences of bigrams starting with $j$, $f_{i|j}$ is more reliable, so a smaller $\lambda$ is used to include more of $f_{i|j}$ in the mixture. Similarly, when there are few occurrences of bigrams starting with $j$, a larger $\lambda$ is used so that the mixture depends more on $f_i$. These different $\lambda$s are set using *deleted interpolation*, which will be described shortly.

---

[1] In general, $F_j = \sum_i F_{i|j}$. For $j = </s>$, $\sum_i F_{i|</s>} = 0$, but it turns out that $F_{</s>}$ is never used.

[2] The method is also sometimes called *deleted interpolation*, after the method used to find $\lambda$. I will use the term "deleted estimation" to refer to the whole process of obtaining a probability distribution using this method, and the term "deleted interpolation" will be reserved for the sub-process of estimating $\lambda$.

[3] There is also a later method called "deleted estimation" by the same authors, which is referred to by Church and Gale (1991).

To summarize the situation so far, we are looking for $Q \equiv \{q_{i|j} \mid i, j = 1, \ldots, W\}$. $q_{i|j}$ depends on the relative frequency $f_i$, the conditional relative frequency $f_{i|j}$, and $\lambda$. $f_i$ and $f_{i|j}$ can be obtained by counting occurrences in the training corpus, and $\lambda$ is about to be determined by deleted interpolation. This gives us all we need to predict the probability of any sentence that uses the vocabulary of the training corpus.

To prepare the reader for the discussion of deleted interpolation, let us start by supposing that there is only one $\lambda$, and show how the value of $\lambda$ can be found using a standard maximum likelihood estimation procedure. Substituting equation 5 into equation 3 yields the following expression for the probability of the training corpus:

$$\Pr(s_1, \ldots, s_n) = \prod_{x=1}^{n} \prod_{y=2}^{len(s_x)} (\lambda f_{w_{x,y}} + (1-\lambda) f_{w_{x,y}|w_{x,y-1}}). \tag{6}$$

To simplify the notation, let

$$u_{x,y} \equiv f_{w_{x,y}} - f_{w_{x,y}|w_{x,y-1}} \tag{7}$$

and

$$v_{x,y} \equiv f_{w_{x,y}|w_{x,y-1}}. \tag{8}$$

So equation 6 can be rewritten

$$\Pr(s_1, \ldots, s_n) = \prod_{x=1}^{n} \prod_{y=2}^{len(s_x)} (\lambda u_{x,y} + v_{x,y}). \tag{9}$$

The value of $\lambda$ that maximizes equation 9 will also maximize the log of equation 9. The log probability is

$$\ell(\lambda) = \sum_{x=1}^{n} \sum_{y=2}^{len(s_x)} \log(\lambda u_{x,y} + v_{x,y}). \tag{10}$$

To find the value of $\lambda$ that maximizes the log probability, we differentiate, set the derivative equal to zero and solve for a root. The first and second derivatives of the log probability are

$$\ell'(\lambda) = \sum_{x=1}^{n} \sum_{y=2}^{len(s_x)} \frac{u_{x,y}}{\lambda u_{x,y} + v_{x,y}} \tag{11}$$

$$\ell''(\lambda) = \sum_{x=1}^{n} \sum_{y=2}^{len(s_x)} -\frac{u_{x,y}^2}{(\lambda u_{x,y} + v_{x,y})^2}. \tag{12}$$

Notice that $\ell''(\lambda) \leq 0$ always. Therefore any root of $\ell'(\lambda) = 0$ will yield a global maximum for $\ell(\lambda)$, subject to checking the endpoints $\lambda = 0$ and $\lambda = 1$. So to find the value of $\lambda$ that maximizes $\Pr(s_1, \ldots, s_n)$, we just find a root, $\hat{\lambda}$, of $\ell'(\lambda) = 0$ using any standard numerical method, such as binary search, and then compare $\Pr(s_1, \ldots, s_n \mid \lambda = \hat{\lambda})$, $\Pr(s_1, \ldots, s_n \mid \lambda = 0)$, and $\Pr(s_1, \ldots, s_n \mid \lambda = 1)$.

However, Jelinek and Mercer (1980) point out that since $\lambda$ is intended to improve the ability of the distribution to predict new data, it must be estimated on data different from that used to calculate $f_i$ and $f_{i|j}$. Deleted interpolation is a method they devised to use all the available training data for estimating both $\lambda$ and $f_i$ and $f_{i|j}$, without actually using the same data for both. They divide the training data into a number, $b$, of blocks, and calculate $b$ different frequency tables, each with a different block of data omitted from the counts. Then $\lambda$ is in effect estimated from the deleted blocks.

To illustrate how deleted interpolation applies to the bigram model, let $f_i^k$ be the relative frequency of the word $i$ calculated with block $k$ of data omitted, and let $f_{i|j}^k$ be the conditional relative frequency of the word $i$ following word $j$, again calculated with block $k$ of the data omitted. Similarly,

$$u_{x,y}^k \equiv f_{w_{x,y}}^k - f_{w_{x,y}|w_{x,y-1}}^k \tag{13}$$

and

$$v_{x,y}^k \equiv f_{w_{x,y}|w_{x,y-1}}^k. \tag{14}$$

The probability of a corpus is

$$\Pr(s_1, \ldots, s_n) = \prod_{k=1}^{b} \prod_{s_x \in k^{th} \text{ block}} \prod_{y=2}^{len(s_x)} (\lambda u_{x,y}^k + v_{x,y}^k). \tag{15}$$

The log probability is

$$\ell(\lambda) = \sum_{k=1}^{b} \sum_{s_x \in k^{th} \text{ block}} \sum_{y=2}^{len(s_x)} \log(\lambda u_{x,y}^k + v_{x,y}^k) \tag{16}$$

Equation 11 is replaced by

$$\ell'(\lambda) = \sum_{k=1}^{b} \sum_{s_x \in k^{th} \text{ block}} \sum_{y=2}^{len(s_x)} \frac{u_{x,y}^k}{\lambda u_{x,y}^k + v_{x,y}^k}, \tag{17}$$

and we can solve for a root of $\ell'(\lambda) = 0$ exactly as before.

With this understanding of how deleted interpolation works for one $\lambda$, let us consider the adjustments needed for the case where several $\lambda$s are used. Suppose we divide the range of possible relative frequencies (the interval $[0, 1]$) into $r$ subintervals, with a separate $\lambda_h$ for each interval $h = 1, \ldots, r$.[4]

---

[4] I do this by dividing the expected range of relative frequencies into $r$ equal subintervals, and then including any

Whenever we calculate the conditional probability for a word $i$ using

$$q_{i|j} = \lambda f_i + (1-\lambda) f_{i|j} \tag{18}$$

we use the $\lambda_h$ for the interval containing $f_j$. When this change is incorporated, the probability of the training data becomes

$$\Pr(s_1, \ldots, s_n) = \prod_{k=1}^{b} \prod_{s_x \in k^{th} \text{ block}} \prod_{h=1}^{r} \prod_{\substack{2 \leq y \leq len(s_x) \text{ and} \\ f_{w_{x,y-1}}^k \in h^{th} \text{ interval}}} (\lambda_h u_{x,y}^k + v_{x,y}^k) \tag{19}$$

The log probability is

$$\ell(\lambda_1, \ldots, \lambda_r) = \sum_{k=1}^{b} \sum_{s \in k^{th} \text{ block}} \sum_{h=1}^{r} \sum_{\substack{2 \leq y \leq len(s_x) \text{ and} \\ f_{w_{x,y-1}}^k \in h^{th} \text{ interval}}} \log(\lambda_h u_{x,y}^k + v_{x,y}^k). \tag{20}$$

The first and second derivatives of the log probability with respect to each $\lambda_h$ are:

$$\frac{\partial}{\partial \lambda_h} \ell(\lambda_1, \ldots, \lambda_r) = \sum_{k=1}^{b} \sum_{s \in k^{th} \text{ block}} \sum_{\substack{2 \leq y \leq len(s_x) \text{ and} \\ f_{w_{x,y-1}}^k \in h^{th} \text{ interval}}} \frac{u_{x,y}^k}{\lambda_h u_{x,y}^k + v_{x,y}^k} \tag{21}$$

$$\frac{\partial^2}{\partial \lambda_h^2} \ell(\lambda_1, \ldots, \lambda_r) = \sum_{k=1}^{b} \sum_{s \in k^{th} \text{ block}} \sum_{\substack{2 \leq y \leq len(s_x) \text{ and} \\ f_{w_{x,y-1}}^k \in h^{th} \text{ interval}}} -\frac{\left(u_{x,y}^k\right)^2}{\left(\lambda_h u_{x,y}^k + v_{x,y}^k\right)^2}. \tag{22}$$

Since $\frac{\partial^2}{\partial \lambda_h^2} \ell(\lambda_1, \ldots, \lambda_r) \leq 0$ for all $h$, it is safe to solve for each $\hat{\lambda}_h$ separately, and then combine the results. This can be done using binary search as before. Deleted estimation requires $O(N)$ computations per iteration, where $N$ is the number of bigrams in the training corpus.

## 2.3  MacKay's Bayesian Smoothing Method

MacKay (1994) notes that "any rational predictive procedure can be made Bayesian", and proposes that by making some assumptions explicit and using Bayesian methods, a better predictive distribution of a similar form can be obtained without any need for cross-validation. Here is how MacKay's method works.

---

higher-than-expected relative frequencies with the $r^{th}$ subinterval.

As before, we are estimating the $W \times W$ probability matrix $Q$, where each element $q_{i|j}$ is the probability that $w_{x,y} = i$ given that $w_{x,y-1} = j$. The Bayesian way to do this is to choose some reasonable prior distribution for $Q$, without looking at the data, and then use Bayes's rule to get a posterior distribution that takes the data into account. MacKay uses Dirichlet priors, mainly because they have the useful property that the posterior has the same form as the prior, thus simplifying the mathematics. Dirichlet priors make use of a *null measure*, which is the vector of relative frequencies that one might expect given no data, and a control parameter, which measures how much spread around the null measure one expects to see in the data. MacKay hypothesizes a model for $Q$ that uses a set of Dirichlet priors, one for each row $\mathbf{q}_{|j}$ of the matrix, all sharing a common null measure, $\mathbf{m}$, and control parameter, $\alpha$, but otherwise independent. He adds an extra degree of freedom to the model for $Q$ by considering $\mathbf{m}$ and $\alpha$ as unknowns to be estimated from the data. He calls this hypothesized model $\mathcal{H}_M$.

Suppose we already know $\mathbf{m}$ and $\alpha$, and let $D \equiv s_1, \ldots, s_x$ represent the training corpus. Then by Bayes's rule,

$$\Pr(Q \mid D, \mathbf{m}, \alpha, \mathcal{H}_M) = \frac{\Pr(D \mid Q, \mathcal{H}_M) \Pr(Q \mid \mathbf{m}, \alpha, \mathcal{H}_M)}{\Pr(D \mid \mathbf{m}, \alpha, \mathcal{H}_M)}. \tag{23}$$

Since by $\mathcal{H}_M$ the rows of $Q$ are independent except for $\mathbf{m}$ and $\alpha$:

$$\Pr(Q \mid \mathbf{m}, \alpha, \mathcal{H}_M) = \prod_j \Pr(\mathbf{q}_{|j} \mid \mathbf{m}, \alpha, \mathcal{H}_M). \tag{24}$$

The Dirichlet priors for the rows of $Q$ have the form

$$\Pr(\mathbf{q}_{|j} \mid \mathbf{m}, \alpha, \mathcal{H}_M) = \frac{1}{Z} \prod_i q_{i|j}^{\alpha m_i - 1} \delta(\sum_i q_{i|j} - 1) \tag{25}$$

where

$$Z = \frac{\prod_i \Gamma(\alpha m_i)}{\Gamma(\sum_i \alpha m_i)}. \tag{26}$$

The likelihood is simply

$$\Pr(D \mid Q, \mathcal{H}_M) = \prod_{s=1}^{x} \prod_{y=1}^{len(s_x)} q_{w_{x,y} \mid w_{x,y-1}} = \prod_j \prod_i q_{i|j}^{F_{i|j}}. \tag{27}$$

The evidence, $\Pr(D \mid \mathbf{m}, \alpha, \mathcal{H}_M)$, does not depend on $Q$, so we can treat it as a constant for now.

It is therefore possible to derive the posterior distribution for $Q$ as follows:

$$\Pr(Q \mid D, \mathbf{m}, \alpha, \mathcal{H}_M) = \frac{\Pr(D \mid Q, \mathcal{H}_M) \prod_j \Pr(\mathbf{q}_{|j} \mid \mathbf{m}, \alpha, \mathcal{H}_M)}{\Pr(D \mid \mathbf{m}, \alpha, \mathcal{H}_M)} \tag{28}$$

$$= \frac{\left(\prod_j \prod_i q_{i|j}^{F_{i|j}}\right) \left(\prod_j \frac{1}{Z_j} \prod_i q_{i|j}^{\alpha m_i - 1} \delta(\sum_i q_{i|j} - 1)\right)}{\Pr(D \mid \mathbf{m}, \alpha, \mathcal{H}_M)} \tag{29}$$

$$= \left( \frac{\prod_j \frac{1}{Z_j}}{\Pr(D \mid \mathbf{m}, \alpha, \mathcal{H}_M)} \right) \prod_j \prod_i q_{i|j}^{F_{i|j} + \alpha m_i - 1} \delta \left( \sum_i q_{i|j} - 1 \right). \qquad (30)$$

Remember that the posterior distributions for the rows of $Q$ are giving us the probabilities of the different values of $q_{i|j}$, considering the data. The predictive probability of $i$ given $j$ is:

$$\Pr(i \mid j, D, \mathbf{m}, \alpha, \mathcal{H}_M) = \int q_{i|j} \Pr(\mathbf{q}_{|j} \mid D, \mathbf{m}, \alpha, \mathcal{H}_M) d^W \mathbf{q}_{|j} \qquad (31)$$

This is just the mean of a Dirichlet distribution, which is known to be

$$\begin{aligned}
\Pr(i \mid j, D, \mathbf{m}, \alpha, \mathcal{H}_M) &= \frac{F_{i|j} + \alpha m_i}{\sum_{i'} \left( F_{i'|j} + \alpha m_{i'} \right)} \\
&= \frac{F_{i|j} + \alpha m_i}{F_j + \alpha}
\end{aligned} \qquad (32)$$

where $F_{i|j}$ and $F_j$ are the counts of bigram $ji$ and token $j$ in the training data as before.

So far we have a formula for $q_{i|j}$ given that we know $\mathbf{m}$ and $\alpha$, but $\mathbf{m}$ and $\alpha$ still need to be estimated from the data. This is also done by Bayesian inference. By Bayes's rule,

$$\Pr(\mathbf{m}, \alpha \mid D, \mathcal{H}_M) = \frac{\Pr(D \mid \mathbf{m}, \alpha, \mathcal{H}_M) \Pr(\mathbf{m}, \alpha \mid \mathcal{H}_M)}{\Pr(D \mid \mathcal{H}_M)}. \qquad (33)$$

The priors for $\alpha$ and $\mathbf{m}$ are uninformative (*i.e.* $\Pr(\log \alpha)$ and $\Pr(\mathbf{m})$ are constant as $\alpha$ and $\mathbf{m}$, respectively, are varied), and independent, and $\Pr(D \mid \mathcal{H}_M)$ is independent of $\mathbf{m}$ and $\alpha$, so we are left with

$$\Pr(\mathbf{m}, \alpha \mid D, \mathcal{H}_M) \propto \Pr(D \mid \mathbf{m}, \alpha, \mathcal{H}_M). \qquad (34)$$

At this point we need an expression for $\Pr(D \mid \mathbf{m}, \alpha, \mathcal{H}_M)$, which previously appeared as a normalizing constant. We know that equation 30 is a product of normalized Dirichlet distributions, each of which has a normalizing constant $\frac{1}{Z'_j}$, where

$$Z'_j = \frac{\prod_i \Gamma \left( F_{i|j} + \alpha m_i \right)}{\Gamma \left( \sum_i F_{i|j} + \alpha m_i \right)}. \qquad (35)$$

So

$$\prod_j \frac{1}{Z'_j} = \frac{\prod_j \frac{1}{Z_j}}{\Pr(D \mid \mathbf{m}, \alpha, \mathcal{H}_M)} \qquad (36)$$

$$\Pr(D \mid \mathbf{m}, \alpha, \mathcal{H}_M) = \prod_j \frac{Z'_j}{Z_j} \qquad (37)$$

$$= \prod_j \left( \frac{\prod_i \Gamma \left( F_{i|j} + \alpha m_i \right)}{\Gamma \left( \sum_i F_{i|j} + \alpha m_i \right)} \frac{\Gamma(\alpha)}{\prod_i \Gamma(\alpha m_i)} \right). \qquad (38)$$

Since $\sum_i m_i = 1$, we can define $u_i \equiv \alpha m_i$ and $\alpha = \sum_i u_i$ to get an equation that depends only on $\mathbf{u}$:

$$\Pr(D \mid \mathbf{u}, \mathcal{H}_M) = \prod_j \left( \frac{\prod_i \Gamma\left(F_{i|j} + u_i\right) \Gamma\left(\sum_i u_i\right)}{\Gamma\left(F_j + \sum_i u_i\right) \prod_i \Gamma\left(u_i\right)} \right). \quad (39)$$

We can recover $\mathbf{m}$ and $\alpha$ from $\mathbf{u}$ at any time by using the above identity.

Strictly, we should integrate out $\mathbf{u}$ when calculating the predictive probabilities, as we did in equation 31. However, under certain conditions that are expected to hold in this case, it is possible to approximate using $\mathbf{u}_{\text{MP}}$, the most probable value of $\mathbf{u}$. That is

$$\begin{align}
\Pr(i \mid j, D, \mathcal{H}_M) &= \int \Pr(\mathbf{u} \mid D, \mathcal{H}_M) \Pr(i \mid j, D, \mathbf{u}, \mathcal{H}_M) d^W \mathbf{u} \quad &(40)\\
&\simeq \Pr(i \mid j, D, \mathbf{u}_{\text{MP}}, \mathcal{H}_M) \quad &(41)\\
&= \frac{F_{i|j} + u_{i\text{MP}}}{F_j + \sum_i u_{i\text{MP}}} \quad &(42)
\end{align}$$

by equation 32.

To find $\mathbf{u}_{\text{MP}}$, we take the log of equation 39, differentiate with respect to $u_i$, set the resulting expression equal to zero, and solve for $u_i$. The digamma function is defined by

$$\Psi(x) \equiv \frac{d}{dx} \log \Gamma(x) \quad (43)$$

and obeys the recurrence relation

$$\Psi(x+1) = \Psi(x) + \frac{1}{x}. \quad (44)$$

For $x > 0.1$ it can be approximated by

$$\Psi(x) \simeq \log(x) - \frac{1}{2x} + O(\frac{1}{x^2}). \quad (45)$$

Since

$$\log \Pr(D \mid \mathbf{u}, \mathcal{H}_M) = \sum_j \left( \sum_i \left(\log \Gamma\left(F_{i|j} + u_i\right) - \log \Gamma\left(u_i\right)\right) + \log \Gamma(\alpha) - \log \Gamma\left(F_j + \alpha\right) \right), \quad (46)$$

then

$$\frac{\partial}{\partial u_i} \log \Pr(D \mid \mathbf{u}, \mathcal{H}_M) = \sum_j \left( \Psi\left(F_{j|i} + u_i\right) - \Psi\left(u_i\right) + \Psi(\alpha) - \Psi\left(F_j + \alpha\right) \right) \quad (47)$$

using the definition of $\Psi(x)$ above and the chain rule. For word bigram models trained on large corpora, it is expected that $u_i < 0.5$ and $\alpha > 1$.[5] Application of recurrence relation 44 yields

$$\Psi(F_{i|j} + u_i) - \Psi(u_i) = \frac{1}{F_{i|j} - 1 + u_i} + \frac{1}{F_{i|j} - 2 + u_i} + \cdots + \frac{1}{2 + u_i} + \frac{1}{1 + u_i} + \frac{1}{u_i} \quad (48)$$

which, for $u_i < 0.5$ can be approximated by

$$\Psi(F_{i|j} + u_i) - \Psi(u_i) \simeq \begin{cases} \frac{1}{u_i} + \sum_{k=1}^{F_{i|j}-1} \frac{1}{k} + \frac{u_i}{k^2} & \text{for } F_{i|j} \geq 1 \\ 0 & \text{for } F_{i|j} = 0. \end{cases} \quad (49)$$

Let $N_{fi}$ be the number of $j$s for which $F_{i|j} \geq f$, and define $G_i \equiv \sum_{f=2}^{\infty} \frac{N_{fi}}{f-1}$ and $H_i \equiv \sum_{f=2}^{\infty} \frac{N_{fi}}{(f-1)^2}$. Then

$$\sum_j \Psi(u_i + F_{i|j}) - \Psi(u_i) \approx \frac{N_{1i}}{u_i} + G_i - u_i H_i. \quad (50)$$

Similarly, using approximation 45,

$$\Psi(F_j + \alpha) - \Psi(\alpha) \approx \log\left(\frac{F_j + \alpha}{\alpha}\right) + \frac{1}{2}\left(\frac{F_j}{\alpha(F_j + \alpha)}\right). \quad (51)$$

Define

$$K(\alpha) \equiv \sum_j \log\left(\frac{F_j + \alpha}{\alpha}\right) + \frac{1}{2}\sum_j \frac{F_j}{\alpha(F_j + \alpha)} \quad (52)$$

so that

$$\sum_j \Psi(F_j + \alpha) - \Psi(\alpha) \approx K(\alpha). \quad (53)$$

Substituting approximations 50 and 53 into equation 47 yields

$$\frac{\partial}{\partial u_i} \log \Pr(D \mid \mathbf{u}, \mathcal{H}_M) \approx \frac{N_{1i}}{u_i} + G_i - u_i H_i - K(\alpha). \quad (54)$$

We set the right hand side of equation 54 equal to zero and solve for $u_i$:

$$u_i = \frac{2N_{1i}}{K(\alpha) - G_i + \sqrt{(K(\alpha) - G_i)^2 + 4H_i N_{1i}}}. \quad (55)$$

Since $G_i$ and $N_{1i}$ do not depend on $\mathbf{u}$ and can be pre-computed, all that remains is to find $\alpha$ such that $\alpha = \sum_i u_i$ and the $u_i$s satisfy equation 55. This can be done by choosing an initial value for $\alpha$, using equation 55 to calculate the $u_i$s, and then iteratively adjusting $\alpha$ until $\alpha = \sum_i u_i$. This method requires $O(W)$ computations per iteration for training.

MacKay expects his Bayesian method to yield superior results over the deleted estimation method

---

[5] Trials with an earlier version of the algorithm presented found an optimal value for $\alpha$ of about 14. So, for a corpus with a 20,000 word vocabulary, the average value of $u_i$ would be expected to be around $\frac{14}{20,000}$.

for two reasons. Equation 32 can be restated as

$$q_{i|j} = \left(\frac{\alpha}{F_j + \alpha}\right) m_i + \left(1 - \frac{\alpha}{F_j + \alpha}\right) f_{i|j} \tag{56}$$

Here, the expression $\frac{\alpha}{F_j+\alpha}$ plays the same role as $\lambda$ in the deleted estimation method, but there is an important difference. While in the deleted estimation method there are one or several $\lambda$s, each set by cross-validation from the training data, here there is effectively a separate $\lambda_j = \frac{\alpha}{F_j+\alpha}$ for each word. These are generated automatically from the frequency of $j$ in the training data, and embody the intuition that the more frequent a word is, the more reliable are the bigram statistics conditioned on that word. So the first advantage is that the individual $\lambda_j$s allow for greater sensitivity in the mixture. The second advantage is that, when a bigram $ji$ does not occur in the training data, the default value of $q_{i|j}$ is $m_i$, a parameter optimized especially for this role, rather than the relative frequency of $i$. This difference would be important in cases where $i$ occurred very frequently following a certain frequent word $k$, but rarely elsewhere. In these cases, the relative frequency would produce a high probability for $ji$, whereas $m_i$ would average the high probability of $ki$ with the low probabilities of other bigrams ending in $i$ and produce a more moderate probability for $ji$.

MacKay suggests an experimental comparison of his method to deleted estimation, but does not carry it out.

## 2.4 Perplexity

The predictive accuracy of competing models is often compared by evaluating the *perplexity* under each model of some common test data.

Perplexity is defined as $2^{H(P,Q)}$, where $H(P,Q)$ is the cross-entropy of the unknown "true" model $P$ as measured by the estimated model $Q$. Brown *et al.* (1992) show that under some reasonable assumptions,

$$H(P,Q) = \lim_{N \to \infty} -\frac{1}{N} \log_2 \Pr_Q(X_1, \ldots, X_N), \tag{57}$$

where $X_1, \ldots, X_N$ is a previously unseen sample of data of size $N$. The better the model $Q$, the lower the cross-entropy $H(P,Q)$, and the smaller the perplexity.

Assuming that, for a large enough test sample,

$$\lim_{N \to \infty} -\frac{1}{N} \log_2 \Pr_Q(X_1, \ldots, X_N) \simeq -\frac{1}{N} \log_2 \Pr_Q(X_1, \ldots, X_N), \tag{58}$$

gives

$$H(P,Q) \simeq -\frac{1}{N} \log_2 \Pr_Q(X_1, \ldots, X_N). \tag{59}$$

Some simple manipulation[6] produces

$$\text{Perplexity} = \Pr\nolimits_Q(X_1, \ldots, X_N)^{-\frac{1}{N}}. \tag{60}$$

But under the bigram models I am using,[7]

$$\Pr\nolimits_Q(X_1, \ldots, X_N) = \prod_{t=2}^{N} q_{X_t|X_{t-1}}, \tag{62}$$

so

$$\text{Perplexity} = \left[\prod_{t=2}^{N} q_{X_t|X_{t-1}}\right]^{-\frac{1}{N}}. \tag{63}$$

So then, if the test sample is large enough, there is a simple method for comparing the accuracy of the competing models: calculate the perplexity of the test sample under each model, and the model with the lowest perplexity wins.

The next section describes an experiment I designed to compare the accuracy of models obtained by using deleted estimation and MacKay's Bayesian methods.

## 3  The Experiment

The text corpus on which the experiment was based was taken from the English portion of Gale and Church's (1991) sentence-aligned version of the Canadian *Hansard*, the proceedings of the Canadian Parliament. This text had already been separated into sentences and stripped of titles, formatting codes, speaker identifiers, and so on (see sample in Appendix B). There are approximately 30 million words of data available in this format: I used about 3 million words of it for this experiment.

Some further pre-processing was required to prepare the data for use by my programs. Sentence numbers were stripped, sentence-begin ($<s>$) and sentence-end ($</s>$) markers were added, and each sentence was placed on a single line. In keeping with the common practice for experiments of this type, punctuation and suffixes beginning with apostrophes were split off from the words they followed, becoming separate tokens. In order to reduce the total number of types in the vocabulary, I also replaced each number by the special token "#".

The resulting sentences were divided into nine blocks of about 1.7 megabytes each, with consecutive sentences going to different blocks. For example, sentences 1, 10, and 19 went to block 1,

---

[6] Shown to me by Peter Brown.

[7] This neat formula is for conceptual purposes. In the implementation, a different manipulation yields

$$\text{Perplexity} = \left(2^{-\frac{1}{N\ln(2)}}\right)^{\sum_{j,i} F_{ji}\ln(q_{i|j})}, \tag{61}$$

which is easier to implement in C.

sentences 2, 11, and 20 to block 2, sentences 3, 12, and 21 to block 3, and so on. This interleaving of the sentences was an important step, because different sections of the text deal with different topics, resulting in different token frequencies. Church and Gale (1991) found that significant differences in language use can occur between different sections of the same corpus, because at different times in the collection period, different topics are being discussed by different people. A model trained on the data from one time period may not fit the data from another time period. By mixing up the sentences, I distributed the data from each time period across all the blocks. The first six blocks were used for training data (about 2 million words), and the test data were extracted from the remaining three blocks. A sample from one of these blocks is shown in Appendix B.

Three versions of the test sample were prepared. First, the smoothing methods being compared only assign probabilities to bigrams composed of tokens that appear in the training data, so they have no way of dealing with previously unseen tokens. Therefore, I removed all sentences that contained a token that did not occur in the training data. This left 14,393 sentences (about 260,000 tokens) in Sample 1. Next, recognizing that the *Hansard* contains many conventional phrases and sentences that might skew the results of the experiment, I removed from Sample 1 all sentences that were duplicated in either the test data or the training data. This left 12,000 sentences (about 243,000 tokens) in Sample 2. Finally, to test whether the sample was large enough for the approximation of perplexity in equation 59 to hold, I pseudo-randomly[8] chose half the sentences in Sample 2 to become Sample 3 (6000 sentences, about 116,000 tokens).

The two smoothing methods have different numbers of parameters to be optimized. For the deleted estimation method, the number of $\lambda$s is set by the experimenter. In MacKay's Bayesian method, there is one $u_i$ for each token in the training data vocabulary. In order to investigate the possible objection that MacKay's Bayesian method would fit the data better simply because of its many parameters, I ran the deleted estimation method with different numbers of $\lambda$s to judge the effect of different numbers of parameters. Also the convergence criteria for each model took number of parameters into account. Within each method, the training continued until *on average* each parameter of the model had converged to eight decimal places.[9] Since some parameters were expected to be very small, it was hoped that eight decimal places would catch at least several significant digits.

The experiment was conducted as follows. First, raw frequencies ($F_j$ and $F_{ji}$) and relative frequencies ($f_j$ and $f_{i|j}$) of tokens and bigrams were obtained from the training data as a whole. Next, the most probable values for the parameters of each model were solved for iteratively. For MacKay's Dirichlet model, this meant solving the simultaneous equations given in section 2.3 to obtain **u**. For deleted estimation, separate frequencies ($f_j^k$ and $f_{i|j}^k$) were first calculated for each block, and then the set of equations given in section 2.2 was solved to obtain $\{\lambda_h\}$. The most probable parameter values for each model were then used to compute probabilities $q_{i|j}$ for each bigram in the test data (i.e., the subset of $Q$ that would actually be used). The result was, in effect, two completely instantiated, competing models of the word sequences in the *Hansard*. Finally, the perplexity of each of the three test data samples was evaluated using each of the models, and the results were compared.

---

[8] The sentences were sorted alphabetically and the first 6000 were taken.
[9] More precisely, on each iteration, the differences for each parameter were summed. If the total was less than 0.000000005 × the number of parameters, convergence was declared.

|  | Model | | | |
|---|---|---|---|---|
|  | deleted estimation | | | Mackay's |
| Sample | 3 $\lambda$s | 15 $\lambda$s | 150 $\lambda$s | |
| Sample 1 |  | 79.60 |  | 79.90 |
| Sample 2 | 89.57 | 88.47 | 88.91 | 89.06 |
| Sample 3 |  | 91.82 |  | 92.28 |

Table 1: Perplexities of the three test data samples under the different models.

The experiment was run on a SPARCserver 1000 with dual "supersparc" CPUs and 128Mb of memory, using a collection of `csh` shell scripts, C programs, and UNIX utilities. In this environment, it took only a few hours of cpu time.

The perplexity of each test sample under each model is given in Table 1.

For all three samples, the perplexities under the deleted estimation model and under MacKay's Bayesian model are nearly the same.

For Sample 2, three deleted estimation models having different numbers of $\lambda$s were tested. The effect of altering the number of $\lambda$s was very small, and of particular interest is the fact that perplexity did not steadily decrease as the number of $\lambda$s increased. This suggests that the large number of parameters in MacKay's Bayesian method does not give it a significant advantage.

Finally, the perplexity results for Sample 3 are close to the corresponding results for Sample 2. This suggests that Sample 2 is large enough to provide a meaningful comparison of models. The fact that the perplexity results for Sample 1 are so much lower than those of Sample 2 probably reflects the high degree of regularity of the extra (conventional) data more than the small increase in test data size.

With regard to resource use, MacKay's algorithm has an advantage. The number of iterations required for each algorithm to converge was comparable. However, a single iteration of MacKay's method requires time linear in the size of the vocabulary, while an iteration of deleted estimation requires time linear in the size of the training corpus. The larger the training corpus, the more significant would be this difference. Also, deleted estimation requires more disk space because it keeps separate count and frequency data for each block of the training corpus.

## 4  Related Work

There are a number of other smoothing methods for estimating probability distributions. In this chapter, I summarize some of these methods and the comparisons that have been made among them. If a method exists in many variations, I have tried to present a basic form applied to bigram probabilities.

One well-known and simple way to handle the problem of zero relative frequencies for bigrams not found in the training corpus is the method of initial counts. Each bigram $ji$ is assigned an initial count of 1 or 0.5 before the training data are examined. The initial counts are added to the observed counts of each bigram in the training data, and the adjusted relative frequency

$$\frac{\text{initial} + F_{ji}}{\sum_{j'i'}(\text{initial} + F_{j'i'})} \qquad (64)$$

is used as the probability of bigram $ji$ in the predictive model. All bigrams with the same observed count have the same probability, and bigrams with an observed count of zero have a small probability due to the initial count. However, Gale and Church (1994) have shown that the method of initial counts gives very poor estimates for bigrams that do occur in the training data.

The Good-Turing method (Good, 1953; Katz, 1987; Church and Gale, 1991) involves not only the observed counts of each bigram in the training corpus but also the number of bigrams that share each count. If $N$ is the total number of bigrams in the corpus, and $N_r$ is the number of bigrams that have observed count r, then the Good-Turing estimate of the probability of a bigram with observed count r is

$$\frac{(r+1)\frac{N_{r+1}}{N_r}}{N}. \qquad (65)$$

Notice that all bigrams with the same count have the same probability, just as with the maximum likelihood estimator or the initial counts method. The probability of each bigram that doesn't appear in the training corpus is

$$\frac{N_1}{N_0 N} \qquad (66)$$

and the probability of all unobserved bigrams adds up to $\frac{N_1}{N}$. In practice, it is usually necessary to smooth the $N_r$s before using them in equation 65. Gale (1994) presents a simple way to smooth these $N_r$s.

Nádas (1984) introduces *parametric empirical Bayes* estimation and compares it with deleted estimation and Good-Turing estimation using perplexity as the measure. Parametric empirical Bayes estimation obtains the probabilities of the bigrams as a result of Bayesian inference. A convenient prior distribution with some unknown parameters is arbitrarily chosen, the posterior probabilities of the bigrams are expressed in terms of the prior, and the parameters of the prior are calculated from the relative frequencies of the bigrams in the training data. The method resembles MacKay's, except that it uses a different prior. Nádas's perplexity figures show that his method is very slightly worse than both Good-Turing and deleted estimation.

Katz's (1987) "back off" method uses a fraction of the corresponding $(n-1)$-gram probability as the probability of an unseen $n$-gram. When applied to bigram models, his method yields the following formula for the conditional probability of a bigram that occurs r times:

$$\Pr(i \mid j) = \begin{cases} \frac{r}{N} & \text{if } r > 5 \text{ (relative frequency)} \\ \frac{r+1}{F_j} \frac{N_{r+1}}{N_r} & \text{if } 0 < r \leq 5 \text{ (Good-Turing)} \\ \gamma \frac{F_i}{N} & \text{if } r = 0 \end{cases} \quad (67)$$

where $F_i$ and $F_j$ are the counts in the training corpus of words $i$ and $j$ respectively, and $\gamma$ is a normalizing constant. Katz claims that his method is time and space efficient, and his perplexity results show that it performs very slightly better than deleted estimation and parametric empirical Bayes.

Church and Gale (1991) compare their "enhanced Good-Turing" and "enhanced deleted estimation" methods to several other standard estimators in a very well thought out and thorough experiment. They first separate the bigrams of the training data into groups depending on the "unigram estimator", which is the product of the unigram relative frequencies of the two constituent words, and then apply another estimation method within each group. Applying Good-Turing within each group (interpreting $N_r$ as the number of bigrams *in this group* that have an observed count of r in the training data) yields the "enhanced Good-Turing" estimates. For the "enhanced deleted estimates", the training data are split into two halves labelled 0 and 1. Then for each group of bigrams above, the following procedure is carried out. Within each half of the training data, count the number of distinct bigrams that occur r times ($N_r^0$ and $N_r^1$) and the total number of occurrences of those bigrams in the other half ($C_r^{01}$ and $C_r^{10}$). Then replace the actual observed count of the bigram by

$$\frac{C_r^{01} + C_r^{10}}{N_r^0 + N_r^1} \quad (68)$$

in calculating the maximum likelihood estimator. For example, in a training corpus of $N$ bigrams, suppose bigram *aa* occurs 300 times in half 0 and 100 times in half 1, bigram *bb* occurs 300 times in half 0 and 400 times in half 1, bigram *cc* occurs 200 times in half 0 and 300 times in half 1, and no other bigrams occur exactly 300 times in either half. $N_{300}^0$ would be 2, $N_{300}^1$ would be 1, $C_{300}^{01}$ would be 500, $C_r^{10}$ would be 200, and the probability of *aa*, which occurs 400 times in the training data,

would be

$$\frac{C_r^{01} + C_r^{10}}{N_r^0 + N_r^1} \frac{1}{N} = \frac{500 + 200}{2 + 1} \frac{1}{N} = \frac{233}{N}. \tag{69}$$

The enhanced Good-Turing estimates and the enhanced deleted estimates are compared with the unigram estimator, the maximum likelihood estimator, and Good-Turing using variances and $t$-scores. The results show that enhanced Good-Turing and enhanced deleted estimation are both better than any of the other methods, but enhanced Good-Turing is more efficient in its use of data. Church and Gale also discuss the need for careful sampling in building the training corpus, and present sophisticated measures for comparing methods.

# 5 Conclusion

I have shown that MacKay's Bayesian model performs as well as deleted estimation in tests on token bigrams, and requires fewer resources. Therefore it is reasonable to recommend that MacKay's Bayesian method replace deleted estimation in future work with token bigrams.

However, most statistical natural language processing systems in the real world use trigrams, not bigrams. Also, Good-Turing estimation seems to be the most accurate of the smoothing methods currently in use. It would be interesting to see whether a Bayesian approach with a different prior model could outperform Good-Turing estimation on trigrams.

# A  Programs

This appendix contains all the programs used to carry out the experiment. "Programs" includes C programs, `csh` scripts, `awk` scripts, and `sed` scripts. The topmost level is given in the form of three diagrams, showing the order and purpose of the programs. Program names are given in bold type. The program listings follow the diagrams, arranged in alphabetical order by name.

## Text Preparation

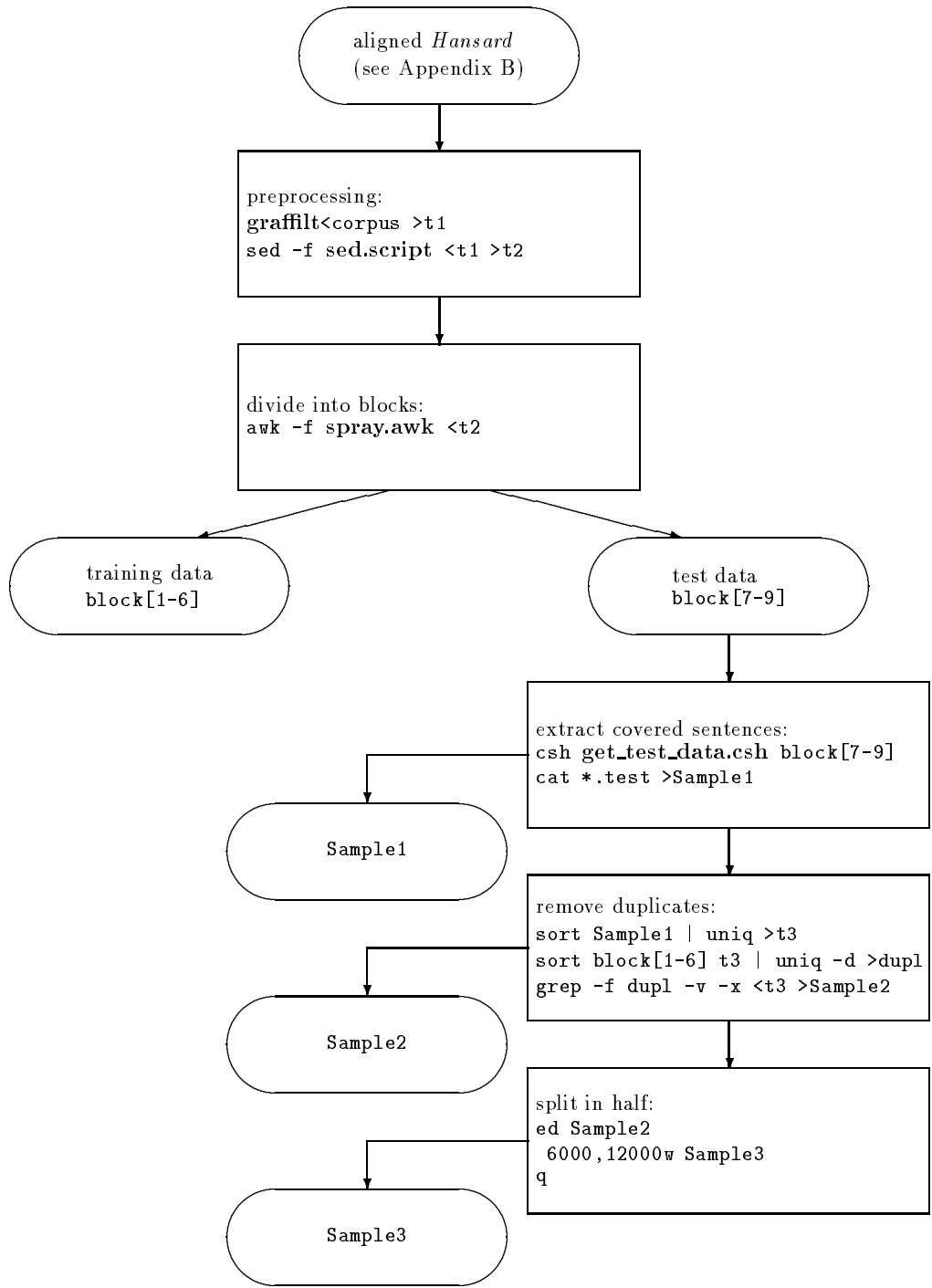

# Constructing the Models

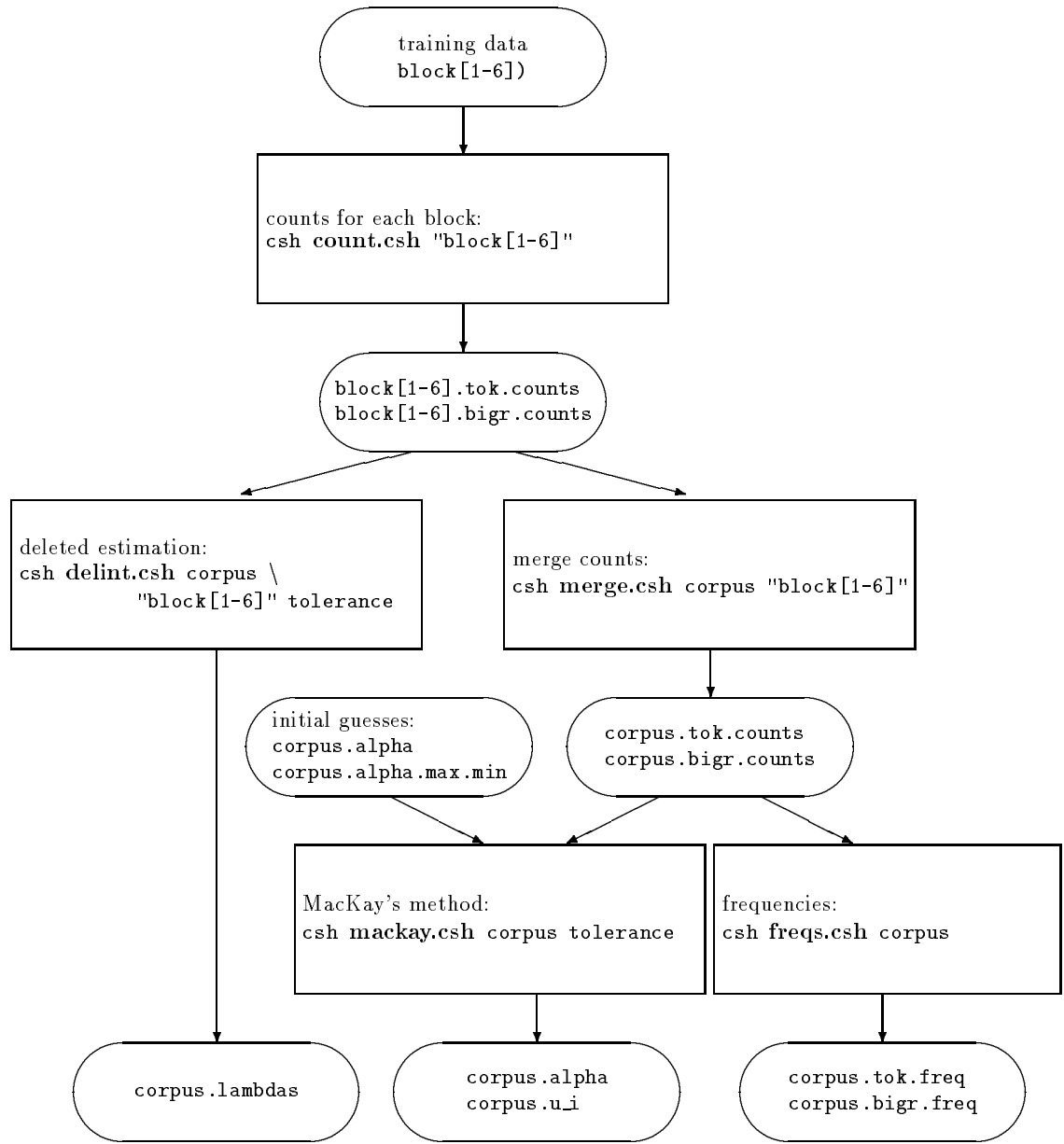

# Testing the Models

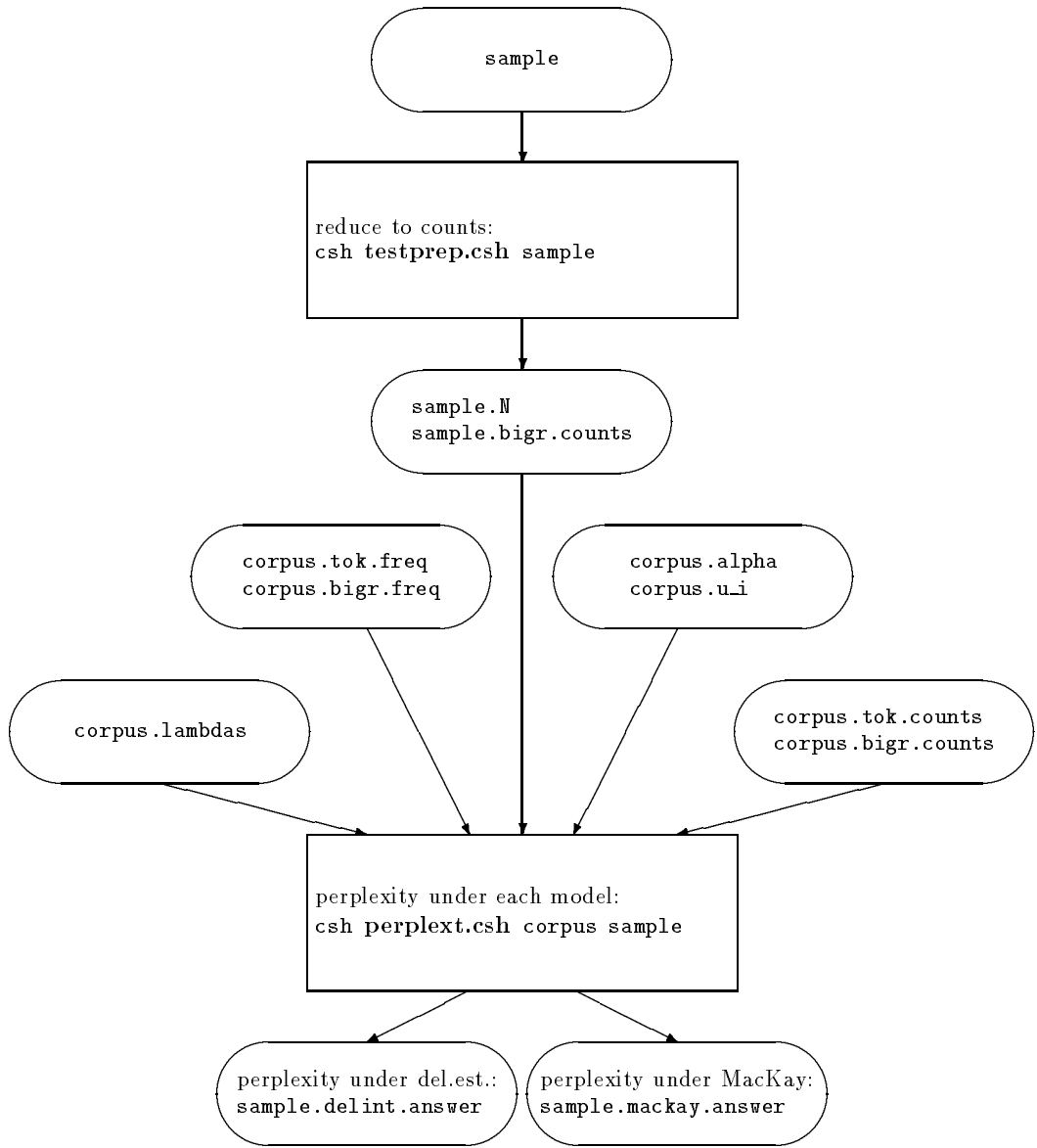

## alpha.c

```c
/* alpha.c */
/*
Increases or decreases alpha depending on sum.u_i.
Input:   $corpus.alpha $corpus.sum.u_i $corpus.alpha.max.min
Output:  alpha on stdout, $corpus.alpha.max.min may be updated.
*/

#include <stdio.h>
#include <math.h>
#define MAXWORDLEN 200

main(int argc, char** argv) {
  FILE* alpha_fp;
  FILE* sum_fp;
  FILE* alpha_max_min_fp;
  double alpha, sum_u_i, alpha_max, alpha_min;

  if (argc!=4) {
    fprintf(stderr,
      "usage: %s alpha sum.u_i alpha.max.min\n",
      argv[0]);
    exit(2);
  };

  /* get alpha */
  if (!(alpha_fp=fopen(argv[1],"r"))) {
    fprintf(stderr, "%s: error opening %s\n",argv[0],argv[1]);
    exit(2);
  };
  fscanf(alpha_fp,"%lg",&alpha);
  fclose(alpha_fp);

  /* get sum_u_i */
  if (!(sum_fp=fopen(argv[2],"r"))) {
    fprintf(stderr, "%s: error opening %s\n",argv[0],argv[2]);
    exit(2);
  };
  fscanf(sum_fp,"%lg",&sum_u_i);
  fclose(sum_fp);

  /* get alpha_max & alpha_min */
  if (!(alpha_max_min_fp=fopen(argv[3],"r"))) {
    fprintf(stderr, "%s: error opening %s\n",argv[0],argv[3]);
    exit(2);
  };
  fscanf(alpha_max_min_fp,"%lg %lg",&alpha_max, &alpha_min);
  fclose(alpha_max_min_fp);

  /* Calculate new alpha */
  if (sum_u_i < alpha) {
    alpha_max=alpha;
    if (alpha_min == -1) /* alpha_min not established yet */
      alpha = alpha / 2;
    else
      alpha = sqrt(alpha_min * alpha_max);
  }
  else if (sum_u_i > alpha) {
    alpha_min=alpha;
```

```c
    if (alpha_max == -1) /* alpha_max not established yet */
      alpha = alpha * 2;
    else
      alpha = sqrt(alpha_min * alpha_max);
  };

  printf("%.15g\n",alpha);

  /* write alpha_max & alpha_min */
  if (!(alpha_max_min_fp=fopen(argv[3],"w"))) {
    fprintf(stderr, "%s: error opening %s for write \n",argv[0],argv[3]);
    exit(2);
  };
  fprintf(alpha_max_min_fp,"%.15g %.15g \n",alpha_max,alpha_min);
  fclose(alpha_max_min_fp);

  return 0;
}
```

## bigrams.c

```c
/* bigrams.c */
/*
Given a sentence of tokens separated by whitespace on stdin,
extracts all token bigrams and writes them one per line on stdout.
*/

#define MAXWORDLENGTH 80

#include <stdio.h>
#include <ctype.h>

main(int argc, char **argv) {
  register int c, state, i, j;
  char buffer[MAXWORDLENGTH];
  state=0;

  while ((c=getchar()) != EOF) {
    switch(state) {
    case 0: /* reading whitespace before first token of sentence */
      if (!isspace(c)) { /* start of first token found */
        state=1;
        i=0;
        buffer[i++]=c;
      }
      break;
    case 1: /* reading and saving first token */
      if (c=='\n') state=0;
      else if (isspace(c)) state=2; /* end of token found */
      else buffer[i++]=c;
      break;
    case 2: /* eating white space between tokens */
      if (c=='\n') state=0;
      else if (!isspace(c)) { /* next token found */
        state=3;
        /* output stored token in buffer[0..i-1] */
        for(j=0; j<i; j++) putchar(buffer[j]);
        putchar('\ ');
        i=0;
```

```c
        /* save and print first char of found token */
        buffer[i++]=c;
        putchar(c);
      };
      break;
    case 3: /* saving and printing a token */
      if (c=='\n') { /* end of sentence */
        state=0;
        putchar('\n');
      }
      else if (isspace(c)) { /* end of token */
        state=2;
        putchar('\n');
      }
      else {
        buffer[i++]=c;
        putchar(c);
      };
      break;
    };
  };
  if (state==3) putchar('\n');
  return 0;
}
```

## bigrfreq.c

```c
/* bigrfreq.c */
/*
Calculates the conditional relative frequency of i|j, given
the total number of bigrams beginning with j and the count of
bigram ji, and prints it on stdout as follows:  freq j i .
*/

#include <stdio.h>
#include <string.h>
#define MAXWORDLEN 200

main(int argc, char **argv) {
  FILE *tokcountfile;  /* total bigrams starting with j */
  FILE *bigrcountfile; /* count j i */

  double tokcount, bigrcount;
  char j[MAXWORDLEN], rest[MAXWORDLEN], token[MAXWORDLEN]="";
  int result;

  if (argc!=3) {
    fprintf(stderr,
            "Usage: bigrfreq corpus.bigr.tots corpus.bigr.counts\n");
    return 1;
  };

  tokcountfile = fopen(argv[1], "r");

  bigrcountfile = fopen(argv[2], "r");
  while ((result=fscanf(bigrcountfile,"%lg %s %s",&bigrcount,j,rest))
         && result!=EOF) {
    while (strcmp(token,j)) {
      if (fscanf(tokcountfile, "%lg %s", &tokcount, token)==EOF) {
```

```
       fprintf(stderr,"error in %s\n",argv[2]);
       exit(2);
     };
    };
    printf("%.15g %s %s\n", bigrcount/tokcount, j, rest);
  };
  fclose(bigrcountfile);
  fclose(tokcountfile);
  return 0;
}
```

## count.csh

```
#count.csh
#Counts tokens and bigrams in a set of text files.
#Input parameter is file name pattern.
#Input files are those files indicated by file name pattern.
#Output files are $file.tok.counts, $file.bigr.counts for each
#input file.

set files=$argv[1]

#Count tokens and bigrams for each input file.
echo Started count.csh
foreach file ($files)
tokens <$file | sort |uniq -c >$file.tok.counts
if ($status) then
  echo tokens failed on $file ; suspend
endif
bigrams <$file |sort |uniq -c >$file.bigr.counts
if ($status) then
  echo bigrams failed on $file ; suspend
endif
end
```

## delint.csh

```
#delint.csh
#Calculation of probabilities by deleted estimation method.
#Parameters are corpus name, file (block) name pattern, tolerance.
#Input files are $file.tok.counts, $file.bigr.counts.
#Output file is $corpus.lambdas.
#count.csh must have already been run, and $corpus.lambdas must
#have been initiallized.

set corpus=$argv[1]
set files=$argv[2]
set tolerance=$argv[3]

set nlambdas=3
@ iterations=100

echo Starting delint.csh

#Make a set of relative frequency files, each omitting one file from
#the counts.
foreach file ($files)
  echo Block $file
```

```
# Create corpus count files, each omitting one file.
  touch $corpus-$file.tok.counts
  touch $corpus-$file.bigr.counts
  foreach otherfile ($files)
#    Merge all count files but $file.
    if ($otherfile != $file) then
      mergehist $corpus-$file.tok.counts $otherfile.tok.counts >$corpus.temp
      if ($status) then
        echo mergehist failed on $otherfile.tok.counts ; suspend
      endif
      mv $corpus.temp $corpus-$file.tok.counts
      mergehist $corpus-$file.bigr.counts $otherfile.bigr.counts >$corpus.temp
      if ($status) then
        echo mergehist failed on $otherfile.bigr.counts ; suspend
      endif
      mv $corpus.temp $corpus-$file.bigr.counts
    endif
  end
# Calculate number of tokens in corpus - file
  awk '$2!="<s>"{sum+=$1}END{print sum}' <$corpus-$file.tok.counts \
      >$corpus-$file.tok.tot
# Calculate relative frequencies ( - file) of tokens and bigrams
  tokfreq $corpus-$file.tok.tot $corpus-$file.tok.counts >$corpus-$file.tok.freq
  if ($status) then
    echo tokfreq failed ; suspend
  endif
  bigrfreq $corpus-$file.tok.counts $corpus-$file.bigr.counts \
      >$corpus-$file.bigr.freq
  if ($status) then
    echo bigrfreq failed ; suspend
  endif
# Make intermediate file
  combine $corpus-$file.bigr.freq $file.bigr.counts |\
      awk '$2!=0{print $1,$3,$4,$2}' >$corpus-$file.f_ji.j.i.F_ji
  sort +1 $corpus-$file.tok.freq $corpus-$file.f_ji.j.i.F_ji | \
      awk 'NF==2{fj=$1}NF==4{print $2,$3,$4,$1,fj}' \
      >$corpus-$file.j.i.F_ji.f_ji.f_j
  sort +1 $corpus-$file.j.i.F_ji.f_ji.f_j $corpus-$file.tok.freq |\
      awk 'NF==2{fi=$1}NF==5{print $1,$2,$3,$4,$5,fi}' \
      >$corpus-$file.j.i.F_ji.f_ji.f_j.f_i
# Delete garbage before proceeding
  rm $corpus-$file.f_ji.j.i.F_ji $corpus-$file.j.i.F_ji.f_ji.f_j
  rm $corpus-$file.bigr.freq $corpus-$file.tok.freq
  rm $corpus-$file.tok.tot $corpus-$file.bigr.counts $corpus-$file.tok.counts
end

#Search for lambda_h's that maximize the probability of the
#training corpus.

#Determine where to look for a zero of l' wrt lambda_h, by
#evaluating the log likelihood (l) and its derivative wrt lambda_h
#(l') for several values of each lambda_h.  These are
#big sums with a contribution from each $file (block).
#My approach is to calculate for each $file and each h, the term
#that file contributes to l and l' given a certain value for
#lambda_h.  Then when all files have been processed, the terms
#are added. This reduces the amount of file access required for
#the calculations.

#Create new accumulator files
```

```
touch $corpus.l_0
touch $corpus.l_1
touch $corpus.lp_0
touch $corpus.lp_.25
touch $corpus.lp_.5
touch $corpus.lp_.75
touch $corpus.lp_1

foreach file ($files)
  terms1 $nlambdas $corpus-$file.j.i.F_ji.f_ji.f_j.f_i $corpus-$file.terms.l_0 \
      $corpus-$file.terms.l_1 $corpus-$file.terms.lp_0 \
      $corpus-$file.terms.lp_.25 $corpus-$file.terms.lp_.5 \
      $corpus-$file.terms.lp_.75 $corpus-$file.terms.lp_1
  if ($status) then
    echo terms1 failed on $file ; suspend
  endif
  mergehist $corpus.l_0 $corpus-$file.terms.l_0 > $corpus.temp
  if ($status) then
    echo mergehist failed on $corpus-$file.terms.l_0 ; suspend
  endif
  mv $corpus.temp $corpus.l_0
  mergehist $corpus.l_1 $corpus-$file.terms.l_1 > $corpus.temp
  if ($status) then
    echo mergehist failed on $corpus-$file.terms.l_1 ; suspend
  endif
  mv $corpus.temp $corpus.l_1
  mergehist $corpus.lp_0 $corpus-$file.terms.lp_0 > $corpus.temp
  if ($status) then
    echo mergehist failed on $corpus-$file.terms.lp_0 ; suspend
  endif
  mv $corpus.temp $corpus.lp_0
  mergehist $corpus.lp_.25 $corpus-$file.terms.lp_.25 > $corpus.temp
  if ($status) then
    echo mergehist failed on $corpus-$file.terms.lp_.25 ; suspend
  endif
  mv $corpus.temp $corpus.lp_.25
  mergehist $corpus.lp_.5 $corpus-$file.terms.lp_.5 > $corpus.temp
  if ($status) then
    echo mergehist failed on $corpus-$file.terms.lp_.5 ; suspend
  endif
  mv $corpus.temp $corpus.lp_.5
  mergehist $corpus.lp_.75 $corpus-$file.terms.lp_.75 > $corpus.temp
  if ($status) then
    echo mergehist failed on $corpus-$file.terms.lp_.75 ; suspend
  endif
  mv $corpus.temp $corpus.lp_.75
  mergehist $corpus.lp_1 $corpus-$file.terms.lp_1 > $corpus.temp
  if ($status) then
    echo mergehist failed on $corpus-$file.terms.lp_1 ; suspend
  endif
  mv $corpus.temp $corpus.lp_1
  rm $corpus-$file.terms.*
end

initlamb $corpus.l_0 $corpus.l_1 $corpus.lp_0 $corpus.lp_.25  $corpus.lp_.5\
    $corpus.lp_.75 $corpus.lp_1 >$corpus.lambdas.work
if ($status) then
  echo initlamb failed ; suspend
endif
rm $corpus.lp_* $corpus.l_*
```

```
#Calculate optimum lambda_h's by binary search.
#Each of the lambda_h's is actually solved for independently,
#but to reduce file access the processes are interleaved.

@ iterid=1
echo Calculating lambda_hs

#For each iteration of binary search
startloop:
echo Iteration $iterid

#The partial derivatives l' wrt lambda_h is a big sum with a
#contribution from each $file (block).  My approach is to
#calculate for each $file, the term it contributes to l' wrt
#lambda_h for each lambda_h, and accumulate them in sum
#accumulation file $corpus.lp.  Then when all $file's have been
#processed, I calculate the new lambdas. This reduces the amount
#of file access required for the calculations.

#Create new sum accumulation files
touch $corpus.lp

#Accumulate sums.
foreach file ($files)
  terms2 $corpus.lambdas.work $corpus-$file.j.i.F_ji.f_ji.f_j.f_i \
      >$corpus-$file.terms.lp
  if ($status) then
    echo terms failed on $file ; suspend
  endif
  mergehist $corpus.lp $corpus-$file.terms.lp > $corpus.temp
  if ($status) then
    echo mergehist failed on $corpus-$file.terms.lp ; suspend
  endif
  mv $corpus.temp $corpus.lp
  rm $corpus-$file.terms.lp
end

#Calculate new lambda_h's.
mv $corpus.lambdas.work $corpus.lambdas.$iterid
lambda $corpus.lambdas.$iterid $corpus.lp > $corpus.lambdas.work
if ($status) then
  echo bracket failed ; suspend
endif
#rm $corpus.lp

#Check if convergence reached
awk '{print $1}' <$corpus.lambdas.$iterid >$corpus.temp1
awk '{print $1}' <$corpus.lambdas.work >$corpus.temp2
diffrnc $corpus.temp1 $corpus.temp2 $tolerance >>$corpus.diffrnc
switch ($status)
  case 0:
    breaksw
  case 1:
    if ($iterid > $iterations) then
      echo Too many iterations ; suspend
    endif
    @ iterid++ ; goto startloop
    breaksw
  case 2:
```

```
     echo diffrnc failed ; suspend
endsw

echo Convergence reached
mv $corpus.temp2 $corpus.lambdas

#Cleanup
rm $corpus-*.j.i.f_ji.f_j.f_i
rm $corpus.diffrnc $corpus.temp1 $corpus.temp2 $corpus-*
```

### diffrnc.c

```
/* diffrnc.c */
/*
Calculates the sum of the differences between corresponding
records of two parallel files of numbers, and checks the total
difference against the tolerance.
Used to compare old and new alpha's and m_i's in Mackay's method,
and to compare old and new lambda's in deleted interpolation method.
Input is oldfile newfile tolerance.
(oldfile and newfile must have the same number of records.)
Output is the total difference on stdout.
Return code is 0 if total difference <= tolerance, 1 if total
difference > tolerance, or 2 for error.
*/

#include <stdio.h>
#include <math.h>
#include <stdlib.h>

main(int argc, char **argv) {
  FILE *oldfile;
  FILE *newfile;
  double old, new, difference=0;

  if (argc!=4) {
    fprintf(stderr,"Usage: diffrnc oldfile newfile tolerance\n");
    exit(2);
  };

  if ((oldfile = fopen(argv[1], "r"))==NULL) {
    fprintf(stderr, "diffrnc:   can't open %s\n",argv[1]);
    exit(2);
  };
  if ((newfile = fopen(argv[2], "r"))==NULL) {
    fprintf(stderr, "diffrnc:   can't open %s\n",argv[2]);
    exit(2);
  };

  while (fscanf(oldfile, "%lg", &old)!=EOF) {
    if (fscanf(newfile, "%lg", &new)==EOF) {
      fprintf(stderr,"diffrnc:   files are different lengths\n");
      exit(2);
    };
    difference += fabs(new - old);
  };
  if (fscanf(newfile, "%lg", &new)!=EOF) {
    fprintf(stderr,"diffrnc:   files are different lengths\n");
    exit(2);
```

```
  };
  printf("%.15g\n",difference);

  if (difference > atof(argv[3])) return 1;
  else return 0;
}
```

## dprobs.c

```
/* dprobs.c */
/*
Given the lambdas, unigram frequencies and bigram frequencies,
it calculates the lambda mixture probabilities for the bigrams
in the test data.
Input is corpus.lambdas, corpus.j.i.f_ji.f_j.f_i (unigram and
bigram frequencies, sorted by j then i, for only those bigrams found
in the test data).
Output is prob (a.k.a. q_ji),j,i , sorted by j then i, on stdout.
*/

#include <stdio.h>
#include <math.h>
#define MAXWORDLEN 200
#define MAXLAMBDAS 200

main(int argc, char* argv[]) {

  FILE *freqs, *lambdas;
  char i[MAXWORDLEN], j[MAXWORDLEN];
  char temp_char[2];
  int r, h;
  double f_ji, f_i, f_j, prob, lambda[MAXLAMBDAS+1], temp_double;
  double interval;

  /* check parameters */
  if (argc!=3) {
    fprintf(stderr,
      "Usage: %s corpus.lambdas corpus.j.i.f_ji.f_j.f_i \n",
      argv[0]);
    exit(2);
  };

  /* load lambdas and calculate interval size*/
  /* For simplicity, h runs from 1 to r rather than 0 to r-1. */
  if ((lambdas=fopen(argv[1], "r")) == NULL){
    fprintf(stderr,"dprobs:  can't open %s\n",argv[1]);
    exit(2);
  };
  r=1;
  while (r<=MAXLAMBDAS &&
    fscanf(lambdas,"%lg %1s",&temp_double,temp_char)!=EOF) {
      lambda[r]=temp_double; r++;};
  if (r>MAXLAMBDAS &&
    fscanf(lambdas,"%lg %1s",&temp_double,temp_char)!=EOF) {
    fprintf(stderr,
      "dprobs:  Too many lambdas.  Change MAXLAMBDAS and recompile.\n" );
    exit(2);
  };
  r--;   /* r now represents the number of lambdas found in the file */
```

```c
  fclose(lambdas);
  interval=0.03/r;  /* size of interval for lambdas 1 through r-1 */
  /* Open files */
  if((freqs = fopen(argv[2],"r")) == NULL){
    fprintf(stderr,"dprobs:  can't open %s\n",argv[2]);
    exit(2);
  };

  /* For each bigram in corpus.j.i.f_ji.f_j.f_i (i.e. each bigram in
     the test data), calculate the probability
  */
  while (fscanf(freqs, "%s %s %lg %lg %lg",
                j, i, &f_ji, &f_j, &f_i)!=EOF) {
    h=floor(f_j/interval + 1);
    if (h>r) h=r;
    prob=lambda[h]*f_i + (1-lambda[h])*f_ji;
    printf("%.15g %s %s\n",prob, j, i);
  };

  fclose(freqs);
  return 0;
}
```

## freqs.csh

```
#freqs.csh
#Calculates relative frequencies of bigrams in a corpus.
#Parameter is corpus name.
#Input files are $corpus.tok.counts, $corpus.bigr.counts.
#Output files are $corpus.bigr.freq, $corpus.tok.freq.

set corpus=$argv[1]

echo Starting freqs.csh
awk '$2!="<s>"{sum+=$1}END{print sum}' <$corpus.tok.counts >$corpus.tok.tot
tokfreq $corpus.tok.tot $corpus.tok.counts >$corpus.tok.freq
if ($status) then
  echo tokfreq failed ; suspend
endif
bigrfreq $corpus.tok.counts $corpus.bigr.counts >$corpus.bigr.freq
if ($status) then
  echo bigrfreq failed ; suspend
endif
rm $corpus.tok.tot
```

## G_i.c

```c
/* G_i.c */
/*
Calculates G_i, H_i and V_i from N_fi.i.f.
Input:  N_fi i f sorted in reverse order by i then f on stdin
  (N_fi is the number of j's for which F_ji = f.)
Output:  G_i H_i V_i i sorted in reverse order by i on stdout.
*/

#include <stdio.h>
#include <string.h>
#include <math.h>
```

```c
#define MAXWORDLEN 200

main(int argc, char** argv) {
  long int f, last_f, f_max_i, last_N_fi, N_fi;
  double G_i, H_i;
  char i[MAXWORDLEN], save_i[MAXWORDLEN]="";
  int result;

  if (argc!=1) {
    fprintf(stderr,"usage: %s \n", argv[0]);
    exit(2);
  };

  f_max_i=0; last_N_fi=0; last_f=0; G_i=0; H_i=0;
  while (1) {
    result=scanf("%ld %s %ld",&N_fi,i,&f);
    /* if i changes or end of file */
    if (result==EOF || strcmp(save_i,i)) {
      /* catch up last i */
      while (last_f>2) {
        last_f--;
        G_i+=last_N_fi/(double)(last_f-1);
        H_i+=last_N_fi/(double)pow(last_f-1,2);
      };
      if (last_f>1) {
        printf("%.15g %.15g %ld %s\n",G_i,H_i,last_N_fi,save_i);
        last_f=1;
      };
      if (result==EOF) break;
      /* start new i */
      G_i=0; H_i=0;
      last_N_fi = N_fi;
      last_f = f_max_i = f;
      strcpy(save_i,i);
      if (f>1) {
        G_i+=N_fi/(double)(f-1);
        H_i+=N_fi/(double)pow(f-1,2);
      }
      else
        printf("%.15g %.15g %ld %s\n",G_i,H_i,N_fi,i);
    }
    /* if i stays the same */
    else {
      /* catch up f */
      while (last_f>f+1) {
        last_f--;
        G_i+=last_N_fi/(double)(last_f-1);
        H_i+=last_N_fi/(double)pow(last_f-1,2);
      };
      /* do this line */
      last_N_fi=N_fi;
      last_f=f;
      if (f>1) {
        G_i+=N_fi/(double)(f-1);
        H_i+=N_fi/(double)pow(f-1,2);
      }
      else
        printf("%.15g %.15g %ld %s\n",G_i,H_i,N_fi,i);
    };
  };
```

```
    return 0;
}
```

### get_test_data.csh

```
set corpus=big
set original=$argv[1]
tokens < $original | sort |uniq -c > $original.tok.counts
combine $corpus.tok.counts $original.tok.counts |\
    awk '$1==0{print $3}' > $corpus.$original.vocab.diff
fgrep -v -f $corpus.$original.vocab.diff $original > $original.test
```

### graffilt.c

```
/*
graffilt.c
Removes embedded newlines from sentences in aligned hansards
obtained from Dave Graff. stdin to stdout
*/

#include<stdio.h>

main () {
  register char char1,char2;

  if ((char1=getchar())==EOF) {
    fprintf(stderr, "graffilt: no input\n");
    exit(2);
  };

  while ((char2=getchar())!=EOF) {
    if (char1!='\n' || char2!=' ') putchar(char1);
    char1=char2;
  };

  putchar(char1);

  return 0;
}
```

### initlamb.c

```
/* initlamb.c */
/*
Determines whether and where to search for a zero of l'.
If there is a zero of l' in the interval [0,1], it produces an
initial guess.  If there is no zero
in the interval, it determines which endpoint maximizes the log
likelihood.
Input files: $corpus.l_0 $corpus.l_1 $corpus.lp_0 $corpus.lp_.25
  $corpus.lp_.5 $corpus.lp_.75 $corpus.lp_1
Output is lambda_h final?(y or n) on stdout.
*/

#include <stdio.h>

main(int argc, char **argv) {
```

```c
FILE *l_0, *l_1, *lp_0, *lp_25, *lp_5, *lp_75, *lp_1;

double l_0_h, l_1_h, lp_0_h, lp_25_h, lp_5_h, lp_75_h, lp_1_h;
int h;

/* Initiallization */
if (argc!=8) {
  fprintf(stderr,"initlamb:  Wrong number of parameters/n");
  exit(2);
};

if ((l_0=fopen(argv[1],"r")) == NULL) {
  fprintf(stderr,"initlamb:  can't open %s\n",argv[1]);
  exit(2);
};
if ((l_1=fopen(argv[2],"r"))==NULL) {
  fprintf(stderr,"initlamb:  can't open %s\n",argv[2]);
  exit(2);
};
if ((lp_0=fopen(argv[3],"r"))==NULL) {
  fprintf(stderr,"initlamb:  can't open %s\n",argv[3]);
  exit(2);
};
if ((lp_25=fopen(argv[4],"r"))==NULL) {
  fprintf(stderr,"initlamb:  can't open %s\n",argv[4]);
  exit(2);
};
if ((lp_5=fopen(argv[5],"r"))==NULL) {
  fprintf(stderr,"initlamb:  can't open %s\n",argv[5]);
  exit(2);
};
if ((lp_75=fopen(argv[6],"r"))==NULL) {
  fprintf(stderr,"initlamb:  can't open %s\n",argv[6]);
  exit(2);
};
if ((lp_1=fopen(argv[7],"r"))==NULL) {
  fprintf(stderr,"initlamb:  can't open %s\n",argv[7]);
  exit(2);
};

/* For each lambda_h, determine whether l' wrt lambda_h has a
   zero in [0,1].  If yes, determine which subinterval it falls
   in and initialize lambda_h.  If no
   determine which endpoint maximizes this lambda_h's
   contribution to the log likelihood.
*/

/* Read the relevant figures for a lambda_h */
while (fscanf(l_0, "%lg %d", &l_0_h, &h)!=EOF) {
  if (fscanf(l_1,"%lg %d",&l_1_h,&h) == EOF) {
    fprintf(stderr, "initlamb:  error in file %s\n",argv[2]);
    exit(2);
  };
  if (fscanf(lp_0,"%lg %d",&lp_0_h,&h) == EOF) {
    fprintf(stderr, "initlamb:  error in file %s\n",argv[3]);
    exit(2);
  };
  if (fscanf(lp_25,"%lg %d",&lp_25_h,&h) == EOF) {
    fprintf(stderr, "initlamb:  error in file %s\n",argv[4]);
    exit(2);
```

```c
    };
    if (fscanf(lp_5,"%lg %d",&lp_5_h,&h) == EOF) {
      fprintf(stderr, "initlamb:  error in file %s\n",argv[5]);
      exit(2);
    };
    if (fscanf(lp_75,"%lg %d",&lp_75_h,&h) == EOF) {
      fprintf(stderr, "initlamb:  error in file %s\n",argv[6]);
      exit(2);
    };
    if (fscanf(lp_1,"%lg %d",&lp_1_h,&h) == EOF) {
      fprintf(stderr, "initlamb:  error in file %s\n",argv[7]);
      exit(2);
    };

fprintf(stderr,
"l_0=%.15g l_1=%.15g lp_0=%.15g lp_25=%.15g lp_5=%.15g lp_75=%.15g lp_1=%.15g\n",
l_0_h,l_1_h,lp_0_h,lp_25_h,lp_5_h,lp_75_h,lp_1_h);

    /*If we happen to have hit the zero already, output it as
      final. */

    if (lp_25_h==0) printf("0.25 y\n");
    else if (lp_5_h==0) printf("0.5 y\n");
    else if (lp_75_h==0) printf("0.75 y\n");

    /*If the signs of l'(0) and l'(1) differ there is a zero
      of l' in [0,1].  Find the subinterval in which it occurs.
    */

    else if (lp_0_h<0 && lp_1_h>0) {
      if (lp_0_h<0 && lp_25_h>0) printf("0.125 n\n");
      else if (lp_25_h<0 && lp_5_h>0) printf("0.375 n\n");
      else if (lp_5_h<0 && lp_75_h>0) printf("0.625 n\n");
      else if (lp_75_h<0 && lp_1_h>0) printf("0.875 n\n");
    }
    else if (lp_0_h>0 && lp_1_h<0) {
      if (lp_0_h>0 && lp_25_h<0) printf("0.125 n\n");
      else if (lp_25_h>0 && lp_5_h<0) printf("0.375 n\n");
      else if (lp_5_h>0 && lp_75_h<0) printf("0.625 n\n");
      else if (lp_75_h>0 && lp_1_h<0) printf("0.875 n\n");
    }

    /* Otherwise choose the endpoint that maximizes the log
       likelihood. */

    else if (l_0_h>l_1_h) printf("0 y\n");
    else printf("1 y\n");
  };

  fclose(l_0);
  fclose(l_1);
  fclose(lp_0);
  fclose(lp_25);
  fclose(lp_5);
  fclose(lp_75);
  fclose(lp_1);

  return 0;
}
```

## K.c

```
/* K.c */
/*
Evaluates K(alpha).
Input:  $corpus.alpha as a file, $corpus.tok.counts on stdin
Output:  K(alpha) on stdout.
*/

#include <stdio.h>
#include <math.h>
#define MAXWORDLEN 200

main(int argc, char** argv) {
  FILE* alpha_fp;
  char j[MAXWORDLEN];
  long int F_j;
  double alpha, sum=0, temp;

  if (argc!=2) {
    fprintf(stderr,"usage: %s alpha < tok.counts \n", argv[0]);
    exit(2);
  };

  /* get alpha */
  if (!(alpha_fp=fopen(argv[1],"r"))) {
    fprintf(stderr, "%s: error opening %s\n",argv[0],argv[1]);
    exit(2);
  };
  fscanf(alpha_fp,"%lg",&alpha);
  fclose(alpha_fp);

  /* evaluate K(alpha) */
  while (scanf("%ld %s",&F_j,j)!=EOF) {
    temp=alpha+F_j;
    sum+=log(temp/alpha) + 0.5 * F_j/(alpha * temp);
  };

  printf("%.15g\n",sum);

  return 0;
}
```

## lambda.c

```
/* lambda.c */
/*
Increases or decreases lambda's depending on $corpus.lp.
Input:  $corpus.lambdas.work $corpus.lp
(lambdas.work has lambda, final?, max, min)
Output:  lambda, final?, max, min on stdout
*/

#include <stdio.h>
#include <math.h>
#define MAXWORDLEN 200

main(int argc, char** argv) {
  FILE* lambdas;
```

```
  FILE* lp;
  double lambda_h, lp_h, max_h, min_h, new_lambda;
  char final[2];
  int h;

  if (argc!=3) {
    fprintf(stderr,
      "usage: %s lambda.work lp\n",
      argv[0]);
    exit(2);
  };

  if ((lambdas=fopen(argv[1],"r")) == NULL) {
    fprintf(stderr,"lambda:  can't open %s\n",argv[1]);
    exit(2);
  };
  if ((lp=fopen(argv[2],"r"))==NULL) {
    fprintf(stderr,"lambda:  can't open %s\n",argv[2]);
    exit(2);
  };

  /* For each lambda_h, if it's not final, calculate the new value
     and print it on stdout, otherwise pass the old value through.
  */
  while (fscanf(lambdas, "%lg %1s", &lambda_h, final)!=EOF) {
    if (fscanf(lp,"%lg %d",&lp_h,&h)==EOF) {
      fprintf(stderr, "lambda:   error on file %s\n",argv[2]);
      exit(2);
    };
    if (final[0]=='n') {
      fscanf(lambdas, "%lg %lg", &max_h, &min_h);
      if (lp_h>0) {
        min_h=lambda_h;
        new_lambda=(max_h+min_h)/2;
        printf("%.15g n %.15g %.15g\n",new_lambda,max_h,min_h);
      }
      else if (lp_h<0) {
        max_h=lambda_h;
        new_lambda=(max_h+min_h)/2;
        printf("%.15g n %.15g %.15g\n",new_lambda,max_h,min_h);
      }
      else printf("%.15g y\n",lambda_h);
    }
    else /* final != 'n' */
      printf("%.15g %s\n",lambda_h, final);
  };

  fclose(lambdas);
  fclose(lp);

  return 0;
}
```

## mackay.csh

```
#mackay.csh
#Estimates alpha and m_i's needed for calculating probabilities
#by MacKays method.
#Command line parameters are corpus name, tolerance.
```

```
#Input files are $corpus.tok.counts, $corpus.bigr.counts,
#$corpus.alpha, and $corpus.alpha.max.min.
#Output files are $corpus.alpha, $corpus.m_i.
#count.csh and merge.csh must be run before this.
#$corpus.alpha $corpus.alpha.max.min must have been externally initialized.

set corpus=$argv[1]
set tolerance=$argv[2]
@ iterid=0
@ iterations=100

echo Starting mackay.csh

#Calculate N_fi's.

awk '{printf "%s %08d\n", $3,$1}' <$corpus.bigr.counts | sort -r | uniq -c |\
  awk '$2!=i{sum=0;i=$2}{sum+=$1;print sum,$2,$3}' >$corpus.N_fi.i.f

#Calculate G_i, H_i and V_i

G_i <$corpus.N_fi.i.f >$corpus.G_i.H_i.V_i

#Repeatedly calculate K(alpha), {u_i} and new alpha values until
#convergence is reached.

echo >$corpus.diffrnc

startloop:

echo Iteration $iterid

#Save old alpha and u_i.
cp $corpus.alpha $corpus.alpha.$iterid
cp $corpus.alpha.max.min $corpus.alpha.max.min.$iterid
if ($iterid > 0) then
  cp $corpus.u_i $corpus.u_i.$iterid
  rm $corpus.neg.flag
endif

#Calculate K(alpha)
K $corpus.alpha <$corpus.tok.counts >$corpus.K
if ($status) then
  echo K failed ; suspend
endif
u_i $corpus.K $corpus.G_i.H_i.V_i $corpus.neg.flag >$corpus.u_i
if ($status) then
  echo u_i failed ; suspend
endif
sumc <$corpus.u_i >$corpus.sum.u_i
alpha $corpus.alpha $corpus.sum.u_i $corpus.alpha.max.min >$corpus.temp
if ($status) then
  echo alpha failed ; suspend
endif
mv $corpus.temp $corpus.alpha

if ($iterid > 0) then
  awk '{print $1}' <$corpus.u_i >$corpus.temp1
  awk '{print $1}' <$corpus.u_i.$iterid >$corpus.temp2
  diffrnc $corpus.temp1 $corpus.temp2 $tolerance >> $corpus.diffrnc
  switch ($status)
```

```
    case 0:
    breaksw
    case 1:
      if ($iterid > $iterations) then
        echo Infinite loop ; suspend
      endif
      @ iterid++ ; goto startloop
    breaksw
    case 2:
      echo diffrnc failed ; suspend
    endsw
else
  @ iterid++ ; goto startloop
endif

echo Convergence reached

#Cleanup
rm $corpus.G_i.H_i.V_i $corpus.sum.* $corpus.N_fi.f.i $corpus.K
rm $corpus.neg.flag  $corpus.alpha.u_i.*  $corpus.alpha.max.min
rm $corpus.temp.*
exit
```

## merge.csh

```
#merge.csh
#Merges the bigram and token counts for the files of a corpus,
#and calculates the total number of tokens.
#Input parameters are corpus name, file name pattern.
#Input files are $file.tok.counts and $file.bigr.counts for each
#file indicated by the file name pattern.
#Output files are $corpus.tok.counts and $corpus.bigr.counts.

set corpus=$argv[1]
set files=$argv[2]

echo Starting merge.csh

#Creates corpus count files if they don't already exist.
touch $corpus.tok.counts
touch $corpus.bigr.counts

#Merge file counts into corpus counts
foreach file ($files)
mergehist $corpus.tok.counts $file.tok.counts >$corpus.temp
if ($status) then
  echo mergehist failed on $file.tok.counts ; suspend
endif
mv $corpus.temp $corpus.tok.counts
mergehist $corpus.bigr.counts $file.bigr.counts >$corpus.temp
if ($status) then
  echo mergehist failed on $file.bigr.counts ; suspend
endif
mv $corpus.temp $corpus.bigr.counts
end
```

# mergehist.c

```c
/* mergehist.c
Merges two sorted histograms.
Adapted from "Tutorial on Text Corpora", Liberman & Marcus, ACL '92.
Handles arbitrary numbers, not just integers.
*/

#include <stdio.h>
#include <stdlib.h>
#include <string.h>
#include <ctype.h>
#define MAXLINE 2048

main(int ac, char** av) {

  char *findval(char*);
  int getline(FILE*, char*, int);
  int putline(char*);

  FILE *infd1, *infd2;
  char in1[MAXLINE], in2[MAXLINE];
  int explain=0;
  register char *p1, *p2;
  register int ret1, ret2;
  register double n1, n2, temp;
  register int needed1, needed2;
  register int c;
  int compare;

  if (ac!=3) explain++;
  else {
    infd1 = fopen(av[1],"r");
    if(infd1 == NULL){
      fprintf(stderr,"can't open %s\n",av[1]);
      explain++;
    };
    infd2 = fopen(av[2],"r");
    if(infd2 == NULL){
      fprintf(stderr,"can't open %s\n",av[2]);
      explain++;
    };
  };
  if (explain){
    fprintf(stderr,"usage: %s hist1 hist2\n", av[0]);
    fprintf(stderr,"\tinput files must be sorted histograms\n");
    exit(2);
  };

  needed1 = needed2 = 1;
  while (1) {
    if (needed1) {
      ret1=getline(infd1,in1,MAXLINE);
      p1=findval(in1);
      n1=atof(in1);
    };
    if (needed2) {
      ret2=getline(infd2,in2,MAXLINE);
      p2=findval(in2);
```

```
      n2=atof(in2);
    };
    if (ret1==0){
      if(ret2==0)exit(0);
      else {
        putline(in2);
        while((c=getc(infd2))!=EOF)putchar(c);
        exit(0);
      };
    }
    else if (ret2==0) {
      putline(in1);
      while((c=getc(infd1))!=EOF)putchar(c);
      exit(0);
    };

    compare = strcmp(p1,p2);
    if (compare==0){
      temp=n1+n2;
      printf("%.15g %s\n",temp,p1);
      needed1 = needed2 = 1;
      continue;
    }
    else if (compare<0) { /* must catch up in first file */
      putline(in1);
      needed1=1;
      needed2=0;
      continue;
    }
    else if (compare>0) { /* must catch up in second file */
      putline(in2);
      needed2=1;
      needed1=0;
      continue;
    };
  };
}

getline(FILE *fd, char *buf, int max) {
  register int c, count=0;
  register char *s=buf;

  while(1) {
    c=getc(fd);
    switch(c) {
    case EOF:
      if (count==0) return(0);
      else {*s = '\0'; return count;};
      break;
    case '\n':
      if (count==0) continue;
      else {*s = '\0'; return count;};
      break;
    default:
      if (++count > max) {
        *s = '\n';
        fprintf(stderr, "overflow: |%s|\n", buf);
        exit(2);
      }
      else *s++ = c;
```

```
      break;
    };
  };
}

char *findval(register char *p) {
  while (*p && isspace(*p)) p++;
  while (*p && !isspace(*p)) p++;
  while (*p && isspace(*p)) p++;
  return(p);
}

putline(register char *p) {
  while (*p) {
    putchar(*p);
    p++;
  };
  putchar('\n');
}
```

## mprobs.c

```
/* mprobs.c */
/*
Given alpha, and a file containing the u_i's and unigram and
bigram frequencies for bigrams occurring in the test data, it
calculates the conditional probabilities for those bigrams, using
the formula
          F_ji  + u_i
   q    = ____________
    i|j   F_j + alpha
where F_ji is the count of bigram ji and F_j is the count
of token j.
Input files are corpus.alpha and corpus.j.i.F_ji.F_j.u_i.
Output is the probability of each bigram in the format prob,j,i
on stdout.
*/

#include <stdio.h>
#define MAXWORDLEN 200

main(int argc, char **argv) {
  FILE *alphafile;
  FILE *j_i_Fji_Fj_ui;

  long int F_ji, F_j;
  double alpha, u_i;
  char j[MAXWORDLEN], i[MAXWORDLEN];
  int result;

  /* Check parameters. */
  if (argc!=3) {
    fprintf(stderr,"Usage: mprobs corpus.alpha corpus.j.i.F_ji.F_j.u_i\n");
    return 2;
  };

  /* Open files. */
  alphafile = fopen(argv[1], "r");
  if (!alphafile) {
```

```c
      fprintf(stderr, "mprobs: cannot open %s\n",argv[1]);
      exit(2);
  };
  result = fscanf(alphafile, "%lg", &alpha);
  if (!result || result == EOF) {
    fprintf(stderr, "%s not an alpha file", argv[1]);
    exit (2);
  };
  fclose(alphafile);

  j_i_Fji_Fj_ui = fopen(argv[2], "r");
  if (!j_i_Fji_Fj_ui) {
    fprintf(stderr, "mprobs: cannot open %s\n",argv[2]);
    exit(2);
  };

  /* Calculate and print probabilities. */
  while (fscanf(j_i_Fji_Fj_ui,"%s %s %ld %ld %lg",j,i,&F_ji,&F_j,&u_i)
         != EOF)
    printf("%.15g %s %s\n",(F_ji + u_i)/(F_j + alpha),j,i);

  fclose(j_i_Fji_Fj_ui);
  return 0;
}
```

## perplex2.c

```c
/* perplex2.c */

/*
Input is a file text.N giving the number bigrams in the test text,
and a table of probabilities and counts for the bigrams in the
test text.
Output is the probability and perplexity of the text on stdout.
*/

#include <stdio.h>
#include <math.h>
#define MAXWORDLEN 200

main(int argc, char **argv) {
  FILE *Nfile, *probs;
  double product=1.0, N, prob;
  char i[MAXWORDLEN],j[MAXWORDLEN];
  int result;
  long int count;

  if (argc!=3) {
    fprintf(stderr,"Usage: perplex2 corpus.N test.something.probs\n");
    return 1;
  };

  Nfile = fopen(argv[1], "r");
  result=fscanf(Nfile, "%lg", &N);
  if (!result || result==EOF) {
    fprintf(stderr, "%s not a N file", argv[1]);
    exit (2);
  };
  fclose(Nfile);
```

```
  probs = fopen(argv[2], "r");
  if (!probs) {
    fprintf(stderr, "perplex2:  can't open %s\n",argv[2]);
    exit(2);
  };

  while ((fscanf(probs,"%ld %lg %s %s",&count,&prob,j,i))!=EOF)
    product*=count*prob;

  fclose(probs);

  printf("Probability=%.15g\tPerplexity=%.15g\n",
    product,product?pow(product,(-1.0/N)):HUGE_VAL);
  return 0;
}
```

### perplext.csh

```
#perplext.csh
#Calculates the probability and perplexity of a test file using
#the two different smoothing methods.
#Input parameters are: corpus name, test file name.
#Input files are:  $corpus.bigr.counts, $corpus.tok.counts,
#$test.bigr.counts, $test.N, $corpus.alpha, $corpus.u_i, $corpus.lambdas,
#$corpus.tok.freq, $corpus.bigr.freq.
#Output files are:   $corpus.delint.answer, $corpus.mackay.answer.

set corpus=$argv[1]
set test=$argv[2]

echo Starting perplext.csh

#Lambda-mixture method.
echo Starting lambda-mixture method
combine $corpus.bigr.freq $test.bigr.counts |\
    awk '$2!=0{print $1,$3,$4}' >$corpus.f_ji.j.i
if ($status) then
  echo first combine failed ; suspend
endif
sort +1 $corpus.tok.freq $corpus.f_ji.j.i | \
    awk 'NF==2{fj=$1}NF==3{print $2,$3,$1,fj}' \
    >$corpus.j.i.f_ji.f_j
if ($status) then
  echo first sort failed ; suspend
endif
sort +1 $corpus.j.i.f_ji.f_j $corpus.tok.freq |\
    awk 'NF==2{fi=$1}NF==4{print $1,$2,$3,$4,fi}' \
    | sort >$corpus.j.i.f_ji.f_j.f_i
if ($status) then
  echo second sort failed ; suspend
endif
dprobs $corpus.lambdas $corpus.j.i.f_ji.f_j.f_i >$test.delint.probs
if ($status) then
  echo dprobs failed ; suspend
endif
combine $test.bigr.counts $test.delint.probs >$corpus.temp
if ($status) then
  echo second combine failed ; suspend
```

```
endif
perplex2 $test.N $corpus.temp>$test.delint.answer
if ($status) then
  echo perplex2 failed ; suspend
endif
rm $corpus.j.i.f_ji.f_j.f_i $corpus.j.i.f_ji.f_j

#Mackay's method.
echo Starting Mackays method
combine $corpus.bigr.counts $test.bigr.counts | \
    awk '$2!=0{print $1,$3,$4}' >$corpus.temp
if ($status) then
  echo first combine failed ; suspend
endif
sort -m +1 $corpus.temp $corpus.tok.counts | \
    awk 'NF==2{Fj=$1}NF==3{print $2,$3,$1,Fj}' >$corpus.j.i.F_ji.F_j
if ($status) then
  echo first sort failed ; suspend
endif
sort +1 $corpus.j.i.F_ji.F_j $corpus.u_i |\
    awk 'NF==2{mi=$1}NF==4{print $1,$2,$3,$4,mi}' |\
    sort >$corpus.j.i.F_ji.F_j.u_i
if ($status) then
  echo second sort failed ; suspend
endif
mprobs $corpus.alpha $corpus.j.i.F_ji.F_j.u_i >$test.mackay.probs
if ($status) then
  echo mprobs failed ; suspend
endif
combine $test.bigr.counts $test.mackay.probs >$corpus.temp
if ($status) then
  echo second combine failed ; suspend
endif
perplex2 $test.N $corpus.temp >$test.mackay.answer
if ($status) then
  echo perplex2 failed ; suspend
endif
rm $corpus.j.i.F_ji.F_j.u_i $corpus.j.i.F_ji.F_j

rm $corpus.temp $corpus.f_ji.j.i
```

sed.script

```
s/^[0-9][0-9]*\.  */<s> /
s/\n / /
s/$/ <\/s>/
s/\\${}/$ /g
s/\\['']\([a-zA-Z]\)/\1/g
s/\\\([&%_]\){}/\1/g
s/\([!?:;]\)/ \1 /g
s/--/ -- /g
s/\.\.\./ ...1 /g
s/Mr\./Mr.1/g
s/Mrs\./Mrs.1/g
s/Dr\./Dr.1/g
s/Ms\./Ms.1/g
s/No\. *\([0-9]\)/No.1 \1/g
s/incl\./incl.1/g
s/\([ap]\)\.m\./\1.m.1/g
```

```
s/\([Hh]\)on\./\1on.1/g
s/\([.,]\) / \1 /g
s/(\(.\)\([^)]\))/( \1\2/g
s/\([^(]\)\(.\))/\1\2 )/g
s/\.\.\.1/.../g
s/Mr.1/Mr./g
s/Mrs.1/Mrs./g
s/Dr.1/Dr./g
s/Ms.1/Ms./g
s/No.1 /No. /g
s/\([Hh]\)on.1/\1on./g
s/incl\.1/incl./g
s/\([ap]\)\.m\.1/\1.m./g
s/"/ " /g
s/'/ ' /g
s/  / /g
```

## spray.awk

```
#Distributes lines of input among the named files
BEGIN{n=1}
n==1{print $0 > "big.1"}
n==2{print $0 > "big.2"}
n==3{print $0 > "big.3"}
n==4{print $0 > "big.4"}
n==5{print $0 > "big.5"}
n==6{print $0 > "big.6"}
n==7{print $0 > "big.7"}
n==8{print $0 > "big.8"}
n==9{print $0 > "big.9"}
{n++; if (n>9) n=1}
```

## sumc.c

```c
/* sumc.c  */
/*
Accepts a sequence of lines on stdin, sums the first field of
each line, and outputs the results on stdout.

Adapted from "Tutorial on Text Corpora", by Liberman and Marcus.
*/

#include <stdio.h>
#include <stdlib.h>
#define MAXLINE 2048

main() {
  register double sum=0;
  char inbuf[MAXLINE];

  while (getline (stdin,inbuf,MAXLINE)) sum += atof(inbuf);
  printf("%.15g\n",sum);

  return 0;
}

getline(FILE *fd, char *buf, int max) {
  register int c, count=0;
```

```
  register char *s = buf;

  while (1) {
    c=getc(fd);
    switch(c) {
    case EOF:
      if (count==0) return (0);
      else {*s = '\0'; return count;};
      break;
    case '\n':
      if (count==0) continue;
      else {*s = '\0'; return count;};
      break;
    default:
      if (++count > max) {
        *s = '\n';
        fprintf(stderr, "overflow: |%s|\n", buf);
        exit(2);
      }
      else *s++ = c;
      break;
    }
  }
}
```

## terms1.c

```
/* terms1.c */
/*
This program calculates the terms contributed by one text block
to each of l (log likelihood) and l' wrt lambda_h, with several
different values of each lambda_h.  These terms are used to
determine the starting points for searching for the lambda_h's
in delint.csh.
Input file:   corpus-file.j.i.F_ji.f_ji.f_j.f_i
Output files: corpus-file.terms.l_0, corpus-file.terms.l_1,
              corpus-file.terms.lp_0,  corpus-file.terms.lp_.25,
              corpus-file.terms.lp_.5, corpus-file.terms.lp_.75,
              corpus-file.terms.lp_1
Usage: terms1 #lambdas corpus-file.j.i.F_ji.f_ji.f_j.f_i corpus-file.terms.l_0
       corpus-file.terms.l_1 corpus-file.terms.lp_0 corpus-file.terms.lp_.25
       corpus-file.terms.lp_.5 corpus-file.terms.lp_.75 corpus-file.terms.lp_1

Note:  +/- FLT_MAX is used to represent +/- Infinity.  This
should be a good enough approximation for our purposes, and it
allows terms of Infinity to be summed using double arithmetic.
*/

#include <stdio.h>
#include <stdlib.h>
#include <math.h>
#include <float.h>
#define MAXWORDLEN 80
#define MAXLAMBDAS 200

main(int argc, char* argv[]) {
  FILE *Fji_fji_fj_fi;
  FILE *l_0, *l_1, *lp_0, *lp_25, *lp_5, *lp_75, *lp_1;
  int r, h;
```

```c
long int F_ji;
double f_ji, f_j, f_i, u_ji;
double suml_0[MAXLAMBDAS+1], suml_1[MAXLAMBDAS+1];
double sumlp_0[MAXLAMBDAS+1], sumlp_25[MAXLAMBDAS+1];
double sumlp_5[MAXLAMBDAS+1], sumlp_75[MAXLAMBDAS+1];
double sumlp_1[MAXLAMBDAS], interval;
char i[MAXWORDLEN], j[MAXWORDLEN];

/* check parameters */
if (argc != 10) {
  fprintf(stderr, "terms1: Wrong number of parameters\n");
  exit(2);
};

/* Calculate size of intervals for lambdas 1 thru r-1.
   Interval r takes up the rest.
   For simplicity, h runs from 1 to r rather than 0 to r-1.
*/
r=atoi(argv[1]);
if (r > MAXLAMBDAS) {
  fprintf(stderr, "terms1:  too many lambdas \n");
  exit(2);
};
interval=0.03/r;

/* initialize temporary sums */
h=1;
while (h<=r) {
  suml_0[h]=suml_1[h]=sumlp_0[h]=sumlp_25[h]=0;
  sumlp_5[h]=sumlp_75[h]=sumlp_1[h]=0;
  h++;
};

/* For each bigram in this block (that is, for each j,i in fji_fj_fi),
calculate the terms that contribute to each sum using the
specified values of lambda_h
*/
if ((Fji_fji_fj_fi=fopen(argv[2],"r"))==NULL) {
  fprintf(stderr,"terms1:  can't open %s\n", argv[2]);
  exit(2);
};

while (fscanf(Fji_fji_fj_fi,"%s %s %ld %lg %lg %lg",
                   j,i,&F_ji,&f_ji,&f_j,&f_i) !=EOF) {
  h=floor(f_j/interval + 1);
  if (h>r) h=r;
  u_ji=f_i - f_ji;
  suml_0[h]+=f_ji?F_ji*log(f_ji):-FLT_MAX;
  suml_1[h]+=F_ji*log(u_ji + f_ji);
  sumlp_0[h]+=f_ji?F_ji*u_ji/f_ji:FLT_MAX;
  sumlp_25[h]+=F_ji*u_ji/(0.25*u_ji + f_ji);
  sumlp_5[h]+=F_ji*u_ji/(0.5*u_ji + f_ji);
  sumlp_75[h]+=F_ji*u_ji/(0.75*u_ji + f_ji);
  sumlp_1[h]+=F_ji*u_ji/(u_ji + f_ji);
  };
fclose(Fji_fji_fj_fi);

/* Write out the new terms */
if ((l_0 = fopen(argv[3],"w"))==NULL) {
  fprintf(stderr,"terms1:  can't open %s\n", argv[3]);
```

```c
      exit(2);
    };
    if ((l_1 = fopen(argv[4],"w"))==NULL) {
      fprintf(stderr,"terms1:  can't open %s\n", argv[4]);
      exit(2);
    };
    if ((lp_0 = fopen(argv[5],"w"))==NULL) {
      fprintf(stderr,"terms1:  can't open %s\n", argv[5]);
      exit(2);
    };
    if ((lp_25 = fopen(argv[6],"w"))==NULL) {
      fprintf(stderr,"terms1:  can't open %s\n", argv[6]);
      exit(2);
    };
    if ((lp_5 = fopen(argv[7],"w"))==NULL) {
      fprintf(stderr,"terms1:  can't open %s\n", argv[7]);
      exit(2);
    };
    if ((lp_75 = fopen(argv[8],"w"))==NULL) {
      fprintf(stderr,"terms1:  can't open %s\n", argv[8]);
      exit(2);
    };
    if ((lp_1 = fopen(argv[9],"w"))==NULL) {
      fprintf(stderr,"terms1:  can't open %s\n", argv[9]);
      exit(2);
    };
    for (h=1;h<=r;h++) {
      fprintf(l_0,"%.15g %d\n",suml_0[h],h);
      fprintf(l_1,"%.15g %d\n",suml_1[h],h);
      fprintf(lp_0,"%.15g %d\n",sumlp_0[h],h);
      fprintf(lp_25,"%.15g %d\n",sumlp_25[h],h);
      fprintf(lp_5,"%.15g %d\n",sumlp_5[h],h);
      fprintf(lp_75,"%.15g %d\n",sumlp_75[h],h);
      fprintf(lp_1,"%.15g %d\n",sumlp_1[h],h);
    };
    fclose(l_0);
    fclose(l_1);
    fclose(lp_0);
    fclose(lp_25);
    fclose(lp_5);
    fclose(lp_75);
    fclose(lp_1);

    return 0;
}
```

### terms2.c

```c
/* terms2.c */
/*
This program calculates the terms contributed by one text block
to each of l' wrt lambda_h and l'' wrt lambda_h, for each
lambda_h.  These terms are used as part of the search for the
lambda_h's, which is carried on in delint.csh.
Input files:  corpus.lambdas, corpus-file.j.i.F_ji.f_ji.f_j.f_i
Output files: corpus-file.terms.lp, corpus-file.terms.lpp
Usage: terms2 corpus.lambdas corpus-file.j.i.F_ji.f_ji.f_j.f_i
            corpus-file.lp corpus-file.lpp
*/
```

```c
#include <stdio.h>
#include <math.h>
#define MAXWORDLEN 80
#define MAXLAMBDAS 200

main(int argc, char* argv[]) {
  FILE *lambdas, *Fji_fji_fj_fi;
  FILE *lp, *lpp;
  int r, h;
  long int F_ji;
  double f_ji, f_j, f_i, u_ji, lambda[MAXLAMBDAS+1], temp_double;
  double sump[MAXLAMBDAS+1], sumpp[MAXLAMBDAS+1], interval;
  char i[MAXWORDLEN], j[MAXWORDLEN], final[MAXLAMBDAS+1], temp_char;

  /* check parameters */
  if (argc != 5) {
    fprintf(stderr, "Usage: terms2 lambdas j.i.F_ji.f_ji.f_j.f_i lp lpp\n");
    exit(2);
  };

  /* Load current values of all the lambdas from lambda file,
     and calculate interval.
     For simplicity, h runs from 1 to r rather than 0 to r-1. */
  if ((lambdas=fopen(argv[1], "r")) == NULL) {
    fprintf(stderr,"terms2:  can't open %s\n", argv[1]);
    exit(2);
  };
  r=1;
  while (r<=MAXLAMBDAS &&
    fscanf(lambdas,"%lg %1s",&lambda[r],&final[r])!=EOF)
    r++;
  if (r>MAXLAMBDAS &&
    fscanf(lambdas,"%lg %1s",&temp_double,&temp_char)!=EOF) {
    fprintf(stderr,
      "terms2:  Too many lambdas.  Change MAXLAMBDAS and recompile.\n" );
    exit(2);
  };
  r--;   /* r now represents the number of lambdas found in the file */
  interval=0.03/r; /* size of intervals for lambdas 1 thru r-1 */

  /* initialize temporary sums */
  h=1;
  while (h<=r) {
    sump[h]=sumpp[h]=0;
    h++;
  };

  /* For each bigram in this block (that is, for each j,i in fji_fj_fi),
  calculate the terms that contribute to l' and l'' using the old lambdas
  */
  if ((Fji_fji_fj_fi=fopen(argv[2],"r"))==NULL) {
    fprintf(stderr,"terms2:  can't open %s\n", argv[2]);
    exit(2);
  };
  while (fscanf(Fji_fji_fj_fi,"%s %s %ld %lg %lg %lg",
                      j,i,&F_ji,&f_ji,&f_j,&f_i) !=EOF) {
    h=floor(f_j/interval + 1);
    if (h>r) h=r;
    u_ji=f_i - f_ji;
```

```
      sump[h]+=F_ji*u_ji/(lambda[h]*u_ji + f_ji);
      sumpp[h]-=F_ji*u_ji*u_ji/pow(lambda[h]*u_ji + f_ji,2);
      };
    fclose(Fji_fji_fj_fi);

    /* Write out the new terms */
    if ((lp = fopen(argv[3],"w"))==NULL) {
      fprintf(stderr,"terms2:  can't open %s\n", argv[3]);
      exit(2);
    };
    if ((lpp = fopen(argv[4],"w"))==NULL) {
      fprintf(stderr,"terms2:  can't open %s\n", argv[4]);
      exit(2);
    };
    for (h=1;h<=r;h++) {
      fprintf(lp,"%.15g %d\n",sump[h],h);
      fprintf(lpp,"%.15g %d\n",sumpp[h],h);
    };
    fclose(lp);
    fclose(lpp);

    return 0;
}
```

## testprep.csh

```
#testprep.csh
#Prepares a test text for the perplexity calculation.
#Parameter is test text filename.
#Output files are $test.bigr.counts, $test.N.

set test=$argv[1]

echo Starting testprep.csh

bigrams <$test | sort |uniq -c >$test.bigr.counts
if ($status) then
  echo bigrams failed ; suspend
endif

sumc <$test.bigr.counts >$test.N
if ($status) then
  echo sumc failed ; suspend
endif
```

## tokens.c

```
/* tokens.c */
/* Given a sentence of tokens separated by whitespace on stdin,
   writes one token per line on stdout.
*/
#include <stdio.h>
#include <ctype.h>

main(int argc, char **argv) {
  register int c, state;
  state=0;
  while ((c=getchar()) != EOF) {
```

```c
      switch(state) {
      case 0: /* reading whitespace */
        if (!isspace(c)) {
           state=1;
           putchar(c);
        }
        break;
      case 1: /* reading a token */
        if (isspace(c)) {
           state=0;
           putchar('\n');
        }
        else putchar(c);
        break;
      };
   };
   if (state==1) putchar('\n');
   return 0;
}
```

## tokfreq.c

```c
/* tokfreq.c */
/*
Given the total number of tokens and the counts for each token,
calculates the relative frequency of each token and prints it to
stdout.
*/

#include <stdio.h>
#define MAXWORDLEN 200

main(int argc, char **argv) {
  FILE *toktotal;
  FILE *tokcounts;

  long int total, count;
  double totald;
  char rest[MAXWORDLEN];
  int result;

  if (argc!=3) {
    fprintf(stderr,"Usage: tokfreq corpus.tok.tot corpus.tok.counts\n");
    return 1;
  };

  toktotal = fopen(argv[1], "r");
  result=fscanf(toktotal, "%ld", &total);
  if (!result || result==EOF) {
    fprintf(stderr, "%s not a total file", argv[1]);
    exit (2);
  };
  fclose(toktotal);

  totald=(double)total;
  tokcounts = fopen(argv[2], "r");
  while ((result=fscanf(tokcounts,"%ld %s",&count,rest))&&result!=EOF)
    printf("%.15g %s\n", count/totald, rest);
  fclose(tokcounts);
```

```
  return 0;
}
```

## u_i.c

```c
/* u_i.c */
/*
Calculates the {u_i} from G_i, H_i, V_i and K.
Input:   $corpus.K $corpus.G_i.H_i.V_i
Output:  $corpus.neg.flag as a file, u_i i on stdout.
*/

#include <stdio.h>
#include <math.h>
#define MAXWORDLEN 200

main(int argc, char** argv) {
  FILE* K_fp;
  FILE* G_i_fp;
  FILE* neg_flag_fp;
  char i[MAXWORDLEN];
  long int V_i;
  double K, G_i, H_i, u_i;
  int neg_flag=0;

  if (argc!=4) {
    fprintf(stderr,"usage: %s K G_i.V_i neg.flag\n", argv[0]);
    exit(2);
  };

  /* get K */
  if (!(K_fp=fopen(argv[1],"r"))) {
    fprintf(stderr, "%s: error opening %s\n",argv[0],argv[1]);
    exit(2);
  };
  fscanf(K_fp,"%lg",&K);
  fclose(K_fp);

  /* open G_i file */
  if (!(G_i_fp=fopen(argv[2],"r"))) {
    fprintf(stderr, "%s: error opening %s\n",argv[0],argv[2]);
    exit(2);
  };

  /* Calculate u_i's */
  while (fscanf(G_i_fp,"%lg %lg %ld %s",&G_i,&H_i,&V_i,i)!=EOF) {
    u_i=(2*V_i)/(K - G_i + sqrt(pow(K-G_i,2) + 4*H_i*V_i) );
    if (u_i<0) neg_flag++;
    printf("%.15g %s\n",u_i,i);
  };
  fclose(G_i_fp);

  if (!(neg_flag_fp=fopen(argv[3],"w"))) {
    fprintf(stderr, "%s: error opening %s\n",argv[0],argv[3]);
    exit(2);
  };
  fprintf(neg_flag_fp,"%d",neg_flag);
  fclose(neg_flag_fp);
```

```
    return 0;
}
```

# B   Data Samples

### Aligned *Hansard* before pre-processing

```
10.  I therefore move, seconded by the hon. member for Cape Breton-The
 Sydney (Mr. Muir):
11.  That the matter of the affront of the Cape Breton Development
 Corporation and its president to the Standing Committee on Regional
 Development, and to two ministers of the Crown, be referred to the
 Standing Committee on Privileges and Elections.
12.  May I say, Mr. Speaker, that I have sent a copy of this to the
 chairman of the committee and to the two ministers involved.
13.  The hon. member has provided the Chair with the required notice of
 his intention to raise this matter today by way of a question of
 privilege.
14.  The hon. member appreciates that what he is proposing now is that
 this motion be given priority over all other business now before the
 House of Commons and that a debate ensue on the suggested question of
 privilege.
15.  I would hesitate very much to suggest to the hon. member that the
 grievance he has should be debated by the House at this time by way of
 a question of privilege.
16.  Perhaps he would allow the Chair to look at the matter within the
 next few hours and to study the notes the hon. member has submitted to
 the Chair for consideration before reaching a decision whether there is
 a prima facie case of privilege that would justify the putting of the
 motion proposed by the hon. member and a debate based on the hon.
 member's motion.
17.  Again I ask that he give the Chair an opportunity to look into the
 matter on his behalf and on behalf of all other members of the House.
18.  1.
19.  For the period April 1, 1973 to January 31, 1974, what amount of
 money was expended on the operation and maintenance of the Prime
 Minister's residence at (a) 24 Sussex Drive, Ottawa (b) Harrington
 Lake, Quebec?
```

### Pre-processed blocks

```
<s> The House met at # p.m. </s>
<s> I therefore move , seconded by the hon. member for Cape Breton-The Sydney ( Mr
. Muir ) : </s>
<s> For the period April # , # to January # , # , what amount of money was expende
d on the operation and maintenance of the Prime Minister 's residence at (a) # Sus
sex Drive , Ottawa (b) Harrington Lake , Quebec ? </s>
<s> Perhaps I did not hear him correctly , but I understood that starred question
No. # was to be answered . </s>
<s> Mr. Speaker , Mr. McInnis had performed legal services in # in respect of a pa
rticular property in the province of Nova Scotia . </s>
<s> # . </s>
<s> With reference to the answer to Question No. # of the First Session of the #th
 Parliament what is the exact purpose of the construction work listed as being carr
ied out in the year #-# at the Prime Minister 's official Ottawa residence and for
 what reasons is such construction deemed necessary ? </s>
<s> # . </s>
<s> The question , therefore , is not one to which the government can respond . </
s>
<s> Under the new proposals presently being considered by the Department of Transp
ort , which provide for a $ # annual landing fee , plus a $ # minimum landing fee
at the international airports of Montreal , Toronto , Winnipeg , Edmonton , Calgar
y and Vancouver , how much revenue is anticipated on an annual basis ? </s>
```

### Bigram conditional relative frequencies

These are the raw conditional relative frequencies as calculated from the training data and stored in $corpus.bigr.freq, so only those bigrams that occurred in the training data are represented. The record layout is $f_{i|j}$ $j$ $i$.

```
0.000333667000333667 's inquiry
0.000333667000333667 's insistence
0.000667334000667334 's institution
0.000333667000333667 's institutions
0.000333667000333667 's instructions
0.00767434100767434 's intention
0.00233566900233567 's intentions
0.00233566900233567 's interest
0.001001001001001 's interests
0.000667334000667334 's interference
0.000333667000333667 's interjection
```

### Bigram probabilities after smoothing with MacKay's method

Only those bigram probabilities needed for the test sample were actually calculated.

```
4.39380132207744e-07 's incompetence
0.000985968776387494 's industrial
9.89359541597536e-07 's initial
0.000327047703278967 's inquest
0.00032997981831265 's inquiry
0.00752128529734649 's intention
0.00228955607194425 's intentions
0.00230075999501251 's interest
0.0016408538264022 's international
0.000327596990651111 's invitation
0.000332481429544508 's job
```

### Bigram probabilities after smoothing with deleted estimation

```
2.14743675590986e-06 's incompetence
0.000799999620882728 's industrial
8.85817661812819e-06 's initial
0.000249866988970725 's inquest
0.000278857385175508 's inquiry
0.00577378370577555 's intention
0.00175349801110414 's intentions
0.00188663908997055 's interest
0.0013034235081431 's international
0.000252685499712857 's invitation
0.000291607790913723 's job
```

# C  Notation

| | |
|---|---|
| $\alpha$ | control parameter in MacKay's method: $\alpha = \sum_i u_i$ |
| $D \equiv s_1, \ldots, s_n$ | the training corpus |
| $F_j$ | number of occurrences of word j in training corpus |
| $F_{i|j}$ | number of occurrences of bigram $ji$ in training corpus |
| $f_i$ | relative frequency of $i$ in training corpus |
| $f_{i|j} \equiv \frac{F_{i|j}}{F_j}$ | conditional relative frequency of bigram $ji$ in training corpus |
| $f_i^k$ | relative frequency of $i$ with block $k$ of data omitted |
| $f_{i|j}^k$ | conditional relative frequency of $ji$ with block $k$ of data omitted |
| $\Gamma(x) \equiv \int_0^\infty u^{x-1} e^{-u} du$ | Gamma function: in general $\Gamma(x+1) = x\Gamma(x)$ |
| $i$ and $j$ | generic words (types) |
| $ji$ | a generic bigram |
| $\lambda, \lambda_h$ | parameters controlling the mix in deleted estimation |
| $\mathbf{m} \equiv \{m_i\}$ | null measure for Dirichlet prior in MacKay's method |
| $N$ | total number of bigrams in a corpus |
| $\Psi(x) \equiv \frac{d}{dx} \log \Gamma(x)$ | digamma function: $\Psi(x+1) = \Psi(x) + \frac{1}{x}$ |
| $Q \equiv \{q_{i|j}\}$ | predictive conditional probability matrix: $q_{i|j} = \Pr(i \mid j)$ |
| $\mathbf{u} \equiv \{u_i\}$ | collapsed parameters in MacKay's method: $u_i = \alpha m_i$ |
| $w_{x,y}$ | $y$th word of $x$th sentence in the training corpus |
| $W$ | number of distinct words in the training corpus vocabulary |